\documentclass[11pt]{article}

\usepackage[final]{acl}

\usepackage{times}
\usepackage{latexsym}

\usepackage[T1]{fontenc}

\usepackage[utf8]{inputenc}

\usepackage{microtype}

\usepackage{inconsolata}

\usepackage{graphicx}
\usepackage{hyperref}
\usepackage{url}
\usepackage{nicefrac}
\usepackage{amsfonts}
\usepackage{bold-extra}
\usepackage{multirow}
\usepackage{multicol}
\usepackage{xspace}
\usepackage[table]{xcolor}
\usepackage{makecell}
\usepackage{array}
\usepackage{booktabs}
\usepackage{amsmath}
\usepackage{tabularx}
\usepackage{dcolumn,caption}
\usepackage{arydshln}
\usepackage{tikz}
\usepackage{pgfplots}
\usepackage{framed}
\usepackage{pifont}
\usepackage{bbm}
\usepackage{enumitem}
\usepackage{amssymb}
\usepackage{adjustbox}
\usepackage{mdframed}
\usepackage{wrapfig}
\usepackage[most,breakable]{tcolorbox}
\usepackage{etoc}
\usepackage{titlesec}
\usepackage{fancyvrb}
\usepackage{listings}
\usepackage{cancel}
\usepackage{setspace}
\usepackage{mathtools}
\usepackage{bm,amsmath,amsfonts}

\usepackage{amsmath}
\usepackage{amssymb}
\usepackage{mathtools}
\usepackage{amsthm}
\usepackage[capitalize,noabbrev]{cleveref}

\definecolor{green}{RGB}{36, 214, 36}
\definecolor{red}{RGB}{235, 30, 30}

\newcommand{\greencheck}{\textcolor{green}{\ding{51}}}
\newcommand{\redcross}{\textcolor{red}{\ding{55}}}

\newcommand{\greenhalfcheck}{\textcolor{green}{\ding{69}}}

\title{TripScore: Aligning LLMs for Real-World Travel Planning via Expert-Calibrated Reward}

\author{
   \textbf{Yincen Qu\textsuperscript{1}}, 
    \textbf{Huan Xiao\textsuperscript{2}},
    \textbf{Feng Li\textsuperscript{1}},
    \textbf{Gregory Li\textsuperscript{3}},
    \textbf{Hui Zhou\textsuperscript{1}},
    \textbf{Xiangying Dai\textsuperscript{1}},
    \textbf{Xuan Huang\textsuperscript{1}},
    \\
    \textsuperscript{1}Trip.com Group,
    \textsuperscript{2}Chongqing University,
    \textsuperscript{3}Stanford University,
    \\
    \small{
    \textbf{Correspondence:} \href{mailto:yc.qu@trip.com}{yc.qu@trip.com}
    }
}

\begin{document}
\maketitle
\begin{abstract}

In our deployed travel-planning service, most users give minimal inputs or free-form requests rather than the structured constraint checklists assumed by existing benchmarks. We therefore present TripScore, a behavior-grounded benchmark and evaluation framework built from real user logs and calibrated against 1,468 pairwise judgments by 203 travel experts. TripScore couples a hierarchical feasibility gate (format and commonsense) with a unified, point-wise reward that aggregates soft quality and preference fulfillment. Using TripScore as both evaluator and reward signal, we benchmark direct prompting, test-time compute, neuro-symbolic solvers, code agents, and fine-tuning. We find that reinforcement learning fine-tuning (e.g., GRPO) provides consistent gains over other approaches under the same base model and practical latency. 

\end{abstract}

\section{Introduction}

Recently, large language models (LLMs)~\citep{gemini,gpt4} have demonstrated promising capabilities in complex reasoning tasks. However, benchmarks such as meeting scheduling and graph coloring~\citep{swe-bench, NaturePlan, graph-coloring} reveal that planning remains a challenging domain to solve comprehensively. Among these challenges, travel planning has gained attention for its real-world impact and intrinsic complexity~\citep{ChinaTravel,travel-agent}. 

Existing travel-planning benchmarks such as TravelPlanner~\citep{TravelPlanner} and ChinaTravel~\citep{ChinaTravel} predominantly treat the task as a hard constraint satisfaction problem, emphasizing feasibility under explicit constraints. However, this framing can drift from authentic usage patterns. We create a web application that enables users to generate travel plans. With explicit user consent, we collected their click behavior and inputs. In our logs analyzing over 300,000 user requests collected over a one-month period, 39.3\% of users specify only minimal constraints (destinations and duration), and 38.3\% submit free-form customized requests. Only 22.4\% tick the detailed constraint checklist assumed by prior work. This gap suggests that progress measured on constraint-heavy benchmarks can misrepresent progress in deployment, where planners must infer intent and trade-offs while still producing executable itineraries.

We therefore introduce \textbf{TripScore}, a behavior-grounded benchmark plus an expert-validated evaluator for real-world travel planning. TripScore is designed not only to measure feasibility and quality, but also to provide an optimizable point-wise reward that can support end-to-end policy optimization. Concretely, TripScore combines a hierarchical feasibility gate (format and commonsense constraints) to prevent invalid plans from being rewarded, and a unified reward that aggregates soft quality and preference satisfaction into a single scalar score. We learn reward weights from pairwise travel-expert preferences and validate robustness via ablation study, making the evaluator stable enough to serve as a training signal rather than merely a post-hoc judge.

\begin{table*}[htbp]
    \centering


\resizebox{\linewidth}{!}{
\small
\begin{tabular}{lcccccccc}
    \toprule
    \multirow{2}{*}{\textbf{Benchmark}} &
    \multirow{2}{*}{\textbf{\# Query}} &
    \multirow{2}{*}{\textbf{\# City}} &
    \multirow{2}{*}{\textbf{Query Type}}  &
    \multirow{2}{*}{\textbf{Query Source}} &
    \textbf{Quality} & 
    \multirow{2}{*}{\textbf{Output Form}}  &
    \textbf{Expert}\\
    \textbf{} &
    \textbf{} &
    \textbf{} &
    \textbf{} &
    \textbf{} &
    \textbf{Evaluation} & 
    \textbf{} &
    \textbf{Validation}
    \\
    \midrule
    TravelPlanner~\citep{TravelPlanner} & 1,225 & 312  & CB Template & Synthetic & \redcross & Pass rate & \redcross \\
    Trip Planning~\citep{NaturePlan} & 1,600 & 48  & CB Template & Synthetic & \redcross & Pass rate & \redcross \\
    ChinaTravel~\citep{ChinaTravel}& 1,154 & 10 & CB Free-Text  & AI-Generated & Rule & Pass rate + Per-dim ranking & \redcross \\
    TripTailor~\citep{TripTailor}& 3,848 & 40  & CB Free-Text &  AI-Generated & LLM & Pass rate + Surpass rate & \redcross \\
    RealTravel~\citep{personalTravel} & 1,155 & 77  & CB Free-Text & AI-Generated   & LLM & Pass rate + Preference rate & \redcross \\
    Travel-Sim~\citep{TravelSim} & 112 & 7  & CB Free-Text & AI-Generated   & Rule + LLM & Pass rate + Per-dim Score & \redcross \\
    \textbf{TripScore-M (ours)}& 4,593 & 821 & CB Template & Real logs (behav.)  & Rule + LLM & Unified score  & \greencheck \\
    \textbf{TripScore-F (ours)}& 277 & 134  & Free-form req. & Real logs (req.) & Rule + LLM & Unified score & \greencheck \\
    \bottomrule
\end{tabular}}

    \caption{Comparison among TripScore and other travel-planning benchmarks. \# Query denotes the number of queries in the dataset; \# City denotes the number of unique cities covered by each benchmark. Quality Evaluation indicates how plan quality is assessed; Output Form denotes the reported metric(s). \greencheck\ indicates expert validation is provided; \redcross\ indicates that quality is not evaluated or expert validation is not provided. Abbreviations: CB = constraints-based, behav. = user click behavior, req. = user free-form request.}
    \label{tab:benchmark_comparison}
    \vspace{-0.5em}
\end{table*}


Our contributions are four-fold. (1) A behavior-grounded dual-track benchmark (TripScore-M for the 39.3\% minimal-constraint users, TripScore-F for the 38.3\% free-form users) curated from real production logs. (2) An expert-validated unified reward whose weights are learned from 1,468 pairwise judgments by 203 travel specialists, reaching 60.29\% agreement with experts (approximately 82.1\% of the noise-adjusted human ceiling) and surpassing LLM-as-judge baselines. (3) A comprehensive method study showing that RL fine-tuning (GRPO) offers the best quality-latency trade-off, narrowing the closed-source model's expert win rate against a 14B open-source model from 0.701 to 0.477 ($p{=}0.628$; a two-sided binomial test does not reject $H_0{:}\,p{=}0.5$). (4) A real-world deployment, where the GRPO-fine-tuned 14B model has fully replaced the prior closed-source LLM endpoint and now serves all user requests in our production travel-planning web application.

\section{Related Work}
Existing travel-planning approaches mix solver-based optimization~\cite{ttg, formal-verify, ChinaTravel, travel-agent} with test-time compute~\cite{hypertree, llm-modullo, TravelSim}. Solvers excel on rigid rule sets but mishandle vague intent, while test-time methods are more flexible yet too slow for deployment. Recent RL-based planners (Planner-R1~\citep{planner-r1}, DeepTravel~\citep{DeepTravel}, TravelSim~\citep{TravelSim}) report gains but rely on synthesized queries and LLM-only rewards without expert validation. On the benchmark side, TravelPlanner~\citep{TravelPlanner} and ChinaTravel~\citep{ChinaTravel} focus on hard-constraint feasibility. TripCraft~\citep{TripCraft}, TP-RAG~\citep{TP-RAG}, TripTailor~\citep{TripTailor}, ITINERA~\citep{ITINERA} and RealTravel~\citep{personalTravel} add quality or preference dimensions but report fragmented per-rule scores, and no prior travel-planning benchmark we are aware of has been validated against domain experts at this scale. We differ on three axes simultaneously, namely \emph{real-user logs} as queries, \emph{expert-calibrated unified reward}, and \emph{end-to-end RL with that reward as signal} (Table~\ref{tab:benchmark_comparison}).

\begin{figure*}[ht!]
    \centering
    \includegraphics[width=0.99\textwidth]{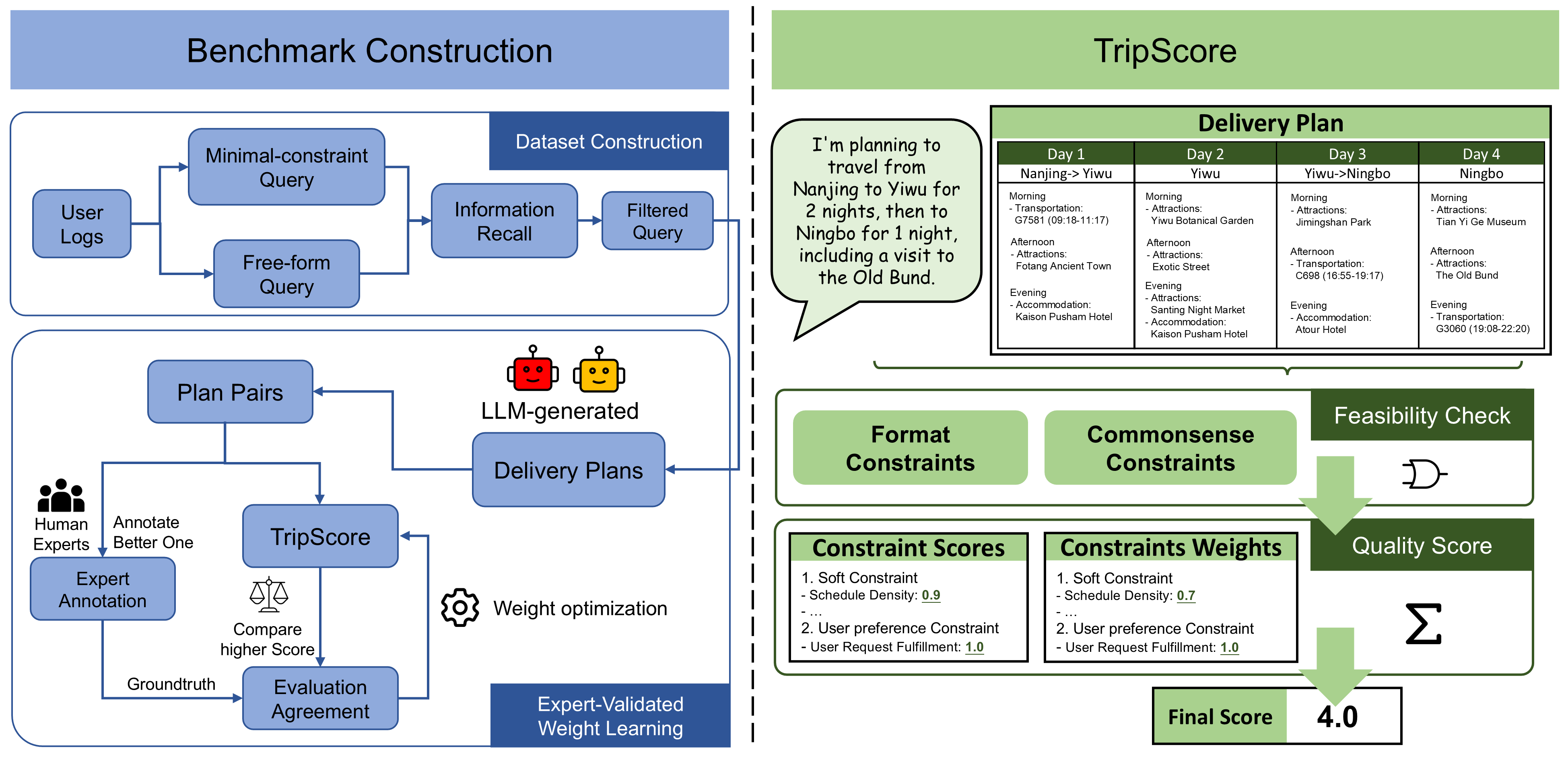}
    \caption{Construction process of TripScore dataset and illustration of TripScore.
    }
    \label{fig:main}
\end{figure*}

\section{The Benchmark and Evaluator}

\textbf{Problem formulation.} Given a user request $q$ (destination, duration, and optional preferences) and a set of candidate resources $\mathcal{C}$ (flights, hotels, attractions from a frozen database), the task is to produce an itinerary $p$ that maximizes the TripScore reward $\mathcal{R}(p;\,q,\mathcal{C})$. The reward decomposes into four dimensions evaluated by distinct mechanisms:

\vspace{-0.5em}
\begin{center}
\small
\setlength{\tabcolsep}{3pt}
\begin{tabular}{lccl}
\toprule
\textbf{Dimension} & \textbf{\#Metrics} & \textbf{Mechanism} & \textbf{Role} \\
\midrule
Format & 4 & Rule & \multirow{2}{*}{Feasibility gate} \\
Commonsense & 4 & Rule &  \\
Soft quality & 7 & Rule + LLM & \multirow{2}{*}{Quality scoring} \\
Preference & 4/1 & Rule / LLM &  \\
\bottomrule
\end{tabular}
\end{center}
\vspace{-0.3em}




\subsection{Dataset Construction}

\textbf{Two tracks from real users.} We deploy a web application (Appendix~\ref{web_application}) that logs roughly 10K daily requests under explicit user consent. \textbf{TripScore-M} samples interactions where users only specify destinations and duration; from logs we retain only these two fields. We then augment each query by sampling structured preferences (budget, pacing, physical effort, attraction interests) from a predefined inventory (Table~\ref{tab:cons-description}, Appendix~\ref{constraints}) and attaching them to the query; these preferences are \emph{not} extracted from user free-text but drawn from the fixed set to enable controlled preference evaluation. \textbf{TripScore-F} retains free-form requests longer than 10 words, spanning specific tasks (e.g., attending a meeting), day-by-day attraction lists (e.g., Day 1: city, Day 2: zoo), and theme trails (e.g., a Harry Potter trail). We additionally filter out requests referencing attractions absent from the frozen resource pool, ensuring all queries have corresponding candidate data. TripScore-F is intentionally smaller (219 test queries) due to per-query expert verification, privacy-filtering cost, and this resource-pool filtering.

\textbf{Information retrieval and quality control.} We bypass the tool-call stage and directly feed candidates retrieved from production-grade flight/hotel/attraction databases (Aug.~2025 frozen snapshot) to isolate planning capability from retrieval errors. Quality control combines empty-retrieval filtering, query-evidence consistency check, and human review of ambiguous or nonsensical requests. The final benchmark totals 4,870 queries (3,493 train / 158 val / 1,219 test) spanning 897 cities, 9,376 hotels, and 10,997 attractions. Full pipeline and detailed statistics are in Appendix~\ref{data_process} and~\ref{data_details}.

\subsection{TripScore Evaluator}

\subsubsection{Constraint Introduction}

TripScore evaluates plans along four dimensions. \textbf{Format} (response structure, information grounding, name/time accuracy, description relevance) and \textbf{Commonsense} (information completeness, location consistency, operating hours, chronological order) together form the feasibility gate. \textbf{Soft constraints} (schedule density, hotel consistency, daytime utilization, geographic clustering, attraction quality) capture general plan quality. We score most with deterministic rules and use LLM ratings only for semantic criteria (iconic landmarks, attraction diversity). \textbf{Personal preferences} are structured for TripScore-M (budget, pacing, physical effort, attraction interests) and free-form for TripScore-F (matched to the user's request via LLM judgment). Full per-metric definitions and Rule/LLM mechanisms are in Table~\ref{tab:cons-description} (Appendix~\ref{constraints}), and cross-benchmark constraints comparison is in Appendix~\ref{constraints_comparison}.

\subsubsection{Two-Stage Evaluation}
We separate feasibility from quality hierarchically. \textbf{Stage 1 (Feasibility Gate)} applies a severe penalty for format violations ($S_{\text{form}}{=}{-}3, S_{\text{com}}=0$) and a non-rewarding score for commonsense violations ($S_{\text{form}}=1, S_{\text{com}}=-1$). Only plans with completely valid formats and commonsense ($S_{\text{form}}=1, S_{\text{com}}=1$) can successfully pass this gate, ensuring that invalid outputs never outscore valid ones. \textbf{Stage 2 (Quality Scoring)} first aggregates sub-metric scores within each quality dimension, then combines the four dimensions into a unified reward:
\begin{align}
\label{aggregate}
\bar{S}_{\text{soft}} &= \frac{\bm{w}_1^\top \bm{S}_{\text{soft}}}{\|\bm{w}_1\|_1}, \quad
\bar{S}_{\text{pref}} = \frac{\bm{w}_2^\top \bm{S}_{\text{pref}}}{\|\bm{w}_2\|_1}, \\
\label{reward}
\mathcal{R}(\bm{S}) &= S_{\text{form}} + S_{\text{com}} + w_3 \bar{S}_{\text{soft}} + w_4 \bar{S}_{\text{pref}}
\end{align}
where $\bm{S}_{\text{soft}}$ and $\bm{S}_{\text{pref}}$ collect sub-scores in $[0,1]$ (seven soft and up to four preference metrics). $\bm{w}_1$ and $\bm{w}_2$ are their within-dimension weights, and $w_3,w_4$ scale the aggregated soft and preference terms. $\bm{\theta}=(\bm{w}_1,\bm{w}_2,w_3,w_4)$ comprises 13 learnable parameters. Sub-metric definitions, feasibility constants, worked examples, and sensitivity analysis are in Appendix~\ref{soft_rule}, Appendix~\ref{preference_rule}, and Appendix~\ref{sensitive_analysis}.

\subsubsection{Expert-Validated Weight Learning}

\textbf{Expert annotation.}
To validate our reward score, we recruited 203 travel experts to rank pairs of travel plans. We restrict annotation to pairs where both plans pass the feasibility gate (format + commonsense constraints), so that expert attention targets subjective quality dimensions rather than rule-verifiable criteria. For each query, we presented two feasible itineraries to three experts and asked them to choose which one is superior. Experts were assigned destinations with which they were familiar. In total, we obtained 1,468 route pairs with three annotations per pair, yielding majority-vote labels; pairs without majority consensus are excluded. The labeled pairs are randomly split at the pair level into train/validation/test for weight learning. Results show moderate inter-annotator agreement (Cohen's $\kappa = 0.5421$ for pairwise; Fleiss's $\kappa = 0.5039$ across three raters), with mean pairwise agreement of 71.69\% and overall three-rater agreement of 59.00\%.
Further details are provided in Appendix~\ref{exper_annotation}.

\textbf{Weight optimization.}
We optimize the weights in Equation~\ref{reward} using expert labels (majority vote per pair). Pairs without a majority consensus (including ties) are excluded from the training set. Given route pairs and their labels, we learn a scoring function $\mathcal{R}(p)$ whose induced pairwise ordering maximizes agreement with expert preferences.

We formulate weight learning as:

\begin{equation}
\bm{\theta}^* = \arg\max_{\bm{\theta}} \sum_{i} \mathbb{I}\big[ \text{sgn}(\Delta \mathcal{R}^{(i)}) = y^{(i)} \big]
\end{equation}
where $y^{(i)}$ is the human label. $\Delta \mathcal{R}^{(i)} = \mathcal{R}(p_1^{(i)}) - \mathcal{R}(p_2^{(i)})$. $\mathcal{R}(p_1^{(i)})$ and $\mathcal{R}(p_2^{(i)})$ are the final reward scores of the plan $p_1$ and $p_2$ correspondingly. 

We perform grid search over 13 parameters, and we choose grid search rather than probabilistic ranking models (e.g., Bradley-Terry) for interpretability of feature importance in this low-dimensional space (Appendix~\ref{weight_optimize}). Empirically, TripScore (grid-search weights) achieves 60.29\% accuracy with Kendall's $\tau{=}0.2026$, versus a Bradley-Terry model trained on the same 13 features at 58.20\% accuracy and $\tau{=}0.1670$ ($p{=}0.003$), confirming that the grid-search approach provides both better expert agreement and auditable, controllable weights. To mitigate overfitting on the 1,468-pair set, we apply nested stratified 5-fold cross-validation, 1{,}000$\times$ bootstrap 95\% CIs, and Kendall's $\tau$ analysis. The selected weights achieve cross-validated validation accuracy $0.6175 \pm 0.0259$ (bootstrap $0.6031$, 95\% CI $[0.5343, 0.6667]$), with positive ordinal alignment ($\tau{=}0.1784$ on val). Under a K-class symmetric-noise model ($K=3$), the 60.29\% raw agreement corresponds to a noise-adjusted latent agreement of $\approx 68.9\%$, covering $\approx 82.1\%$ of the human reliability ceiling ($r \approx 83.9\%$). The details are in Appendix~\ref{noise_adjusted}.

\textbf{Evaluation agreement.} Across LLM backbones, TripScore surpasses the best LLM-as-judge baselines~\citep{llm-judge,llm-judge-travel} on validation agreement. In Table~\ref{tab:evaluation_experiment}, we see 60.29\% with Gemini-3-flash and 61.22\% with Gemini-3.1-pro. Point-wise judges are unstable to ties (Gemini-3-flash drops 11.1\,pp once ties count as wrong). Pair-wise judging is competitive but requires $O(n\log n)$ comparisons. Our point-wise score avoids both, with full tie-handling definitions and per-backbone results. The details are provided in Appendix~\ref{evaluation_accuracy}.

\begin{table}[htbp]

    \centering
    

    
          



\resizebox{0.45\textwidth}{!}{
\begin{tabular}{llcc}
    \toprule
    \textbf{Method} &
    \textbf{LLM} &
    \textbf{Accuracy} &
    \textbf{Kendall's $\tau$} 

    \\
    
    \midrule
          \multirow{5}{*}{Point-wise}  & GPT-5.4 & 47.42 (57.06) & 0.0915  \\
           & Claude-opus-4.6 & 43.19 (62.04) &  0.1745 \\
           & Gemini-3.1-pro & 51.17 (61.24) & 0.1890  \\
           & Gemini-3-flash & 51.64 (60.77) & 0.1701  \\
           & DeepSeek-V3.2 & 51.43 (54.55) & 0.0618 \\
          \midrule
          
     \multirow{5}{*}{Pair-wise} & GPT-5.4 & 54.46 & 0.0820  \\
      & Claude-opus-4.6 & 57.75 & 0.1575 \\
      & Gemini-3.1-pro & 60.09 & 0.1977 \\
      & Gemini-3-flash & 58.22 & 0.1590 \\
      & DeepSeek-V3.2 & 54.46 & 0.0811 \\

         \midrule

       \multirow{4}{*}{Ours} 
       & Gemini-3.1-pro &  61.22 & 0.2155 \\
       & Gemini-3-flash & 60.29 & 0.2026 \\
       & GPT-5.4 & 56.50 & 0.1307 \\
       & DeepSeek-V3.2 & 57.84 & 0.1573 \\

    \bottomrule
\end{tabular}}

    \caption{Comparison with human annotations for our method and LLM-as-judge baselines. All accuracy values are computed on pairs with a clear majority-vote winner (tie/neither cases excluded from the denominator). For point-wise methods, values in parentheses further exclude cases where the judge itself returns a tie.
    }
    \label{tab:evaluation_experiment}
\end{table}

\begin{table*}[t]
\centering

\resizebox{0.98\textwidth}{!}{
\small

\begin{tabular}{llcccccccccc}
\toprule
 \multirow{2}{*}{\textbf{Method}}  & \multirow{2}{*}{\textbf{LLM}} & \multicolumn{4}{c}{\textbf{TripScore-M}} & \multicolumn{4}{c}{\textbf{TripScore-F}} \\
 
\cmidrule(lr){3-6} \cmidrule(lr){7-10}

  & & \small FPR$\uparrow$ & \small CPR $\uparrow$ & \small Reward$\uparrow$  & \small Time$\downarrow$ & \small FPR$\uparrow$ & \small CPR$\uparrow$ & \small Reward$\uparrow$ & \small Time$\downarrow$  \\
  
\midrule
\multicolumn{12}{c}{\textbf{Direct}}           \\ 
\midrule

\multirow{7}{*}{Direct} & GPT-5.4 & 82.18 & 92.02 & 1.69 & 25.09 & 89.35 & 78.76 & 1.81 & 27.70 \\
  & Claude Opus 4.6 & 79.33 & 90.49 & 1.49 & 36.86 & 72.20 & 85.85 & 1.49 & 44.04 \\
  & Kimi-K2.5 & 78.50 & 91.97 & 1.46 & 39.84 & 66.66 & 80.56 & 1.12 & 39.51 \\
  & DeepSeek-V3.2 & 73.92 & 96.00 & 1.26 & 27.61 & 75.78 & 78.10 & 1.44 & 25.03 \\
  & Qwen3-8B & 27.14 & 94.98 & -1.31 & 9.37 & 39.73 & 56.30 & -1.00 & 10.48 \\
  & Qwen3-14B & 39.93 & 88.70  & -0.72 &  12.35 &  51.59 & 67.26  & -0.24 & 13.15 \\
  & Qwen3-32B & 42.32 & 95.13 & -0.53 & 19.33 & 60.27 & 71.86 & 0.49 & 19.83 \\
\midrule
\multicolumn{10}{c}{\textbf{Test-Time Compute}}           \\ 
\midrule

\multirow{3}{*}{CoT}   & GPT-5.4 & 83.83 & 89.43  & 1.72 & 29.48 & 82.19 & 80.55 & 1.92 & 27.14 \\
  & Qwen3-8B & 22.53 & 93.47  & -1.61 & 13.77 & 36.07 & 51.89 & -1.24 & 13.51 \\
  & Qwen3-14B  & 30.85 & 83.43 & -1.33 & 17.59 & 45.79 & 70.40  & -0.44 & 17.27 \\

\cmidrule{1-10}

\multirow{3}{*}{ReAct} & GPT-5.4 & 81.77 & 91.68 & 1.66 & \cellcolor{gray!25}{42.87} & 87.67 & 76.55 & 2.16 & \cellcolor{gray!25}{43.93} \\
  & Qwen3-8B & 31.10 & 93.14  & -1.23 & 18.98 & 40.10 & 72.68  & -0.73 & 21.88 \\
  & Qwen3-14B  & 54.89 & 81.79 & -0.06 & 26.16 & 60.73 & 80.45  & 0.66 & 31.75 \\

\cmidrule{1-10}

  \multirow{3}{*}{LLM-Modulo}& GPT-5.4 & \textbf{93.82} & 91.09 & \textbf{2.36} & \cellcolor{gray!25}{102.7} & \textbf{94.95} & \underline{86.47} & \textbf{3.13} & \cellcolor{gray!25}{106.7} \\
  & Qwen3-8B & 44.33 & 90.32 & -0.49 & 34.17 & 50.22 & 72.72  & -0.16 & \cellcolor{gray!25}{40.23} \\ 
  & Qwen3-14B  & 49.54 & 85.42 & -0.29 & \cellcolor{gray!25}{64.38}  & 56.62 & 81.56 & 0.27 & \cellcolor{gray!25}{54.14} \\

\cmidrule{1-10}

  \multirow{3}{*}{HyperTree}  & GPT-5.4 & 78.26 & 91.72  & 1.44 & \cellcolor{gray!25}{58.10} & 89.95 & 79.69  & 2.03 & \cellcolor{gray!25}{45.96} \\
 & Qwen3-8B & 14.67 & 85.75 & -2.19 & 31.31 & 4.56 & \textbf{100.00} & -2.62 & 35.92 \\
 & Qwen3-14B   & 27.25 & 66.92 & -1.66 & 34.21 & 42.01 & 63.04 & -0.72 & \cellcolor{gray!25}{41.71} \\
    
\midrule
\multicolumn{10}{c}{\textbf{Neural-Symbolic}}           \\ 
\midrule

TTG & GPT-5.4 & 37.10 & \textbf{99.38} & -0.56 & 2.21  & 12.78 & \textbf{100.00}  & -2.17 & 2.03 \\ %

\cmidrule{1-10}

\multirow{2}{*}{NESY} & GPT-5.4 & 57.96 & \underline{98.74}  & 0.60 & \cellcolor{gray!25}{77.92}& 1.82 & \textbf{100.00} & -2.89 & \cellcolor{gray!25}{46.44} \\  

& Qwen3-8B & 47.37 & 98.04  & -0.25 & \cellcolor{gray!25}{62.58} & 2.73 & \textbf{100.00} & -2.83 & \cellcolor{gray!25}{42.12} \\ 

\midrule
\multicolumn{10}{c}{\textbf{Fine-Tuning}}           \\ 
\midrule

 \multirow{2}{*}{SFT} & Qwen3-8B & 74.02 & 91.23  & 1.17 & 9.00 & 77.16 & 82.84 & 1.53 & 11.85 \\
  & Qwen3-14B & 82.80 &  93.41   & 1.70  & 12.69 & 85.38 &  76.47 & 1.89  & 15.34  \\

\cmidrule{1-10}

   \multirow{2}{*}{RFT} & Qwen3-8B & 74.49 & 93.66 &  1.26 & 9.77 & 82.19 & 81.11 & 1.82 & 12.21  \\
    & Qwen3-14B  & 84.75 &  93.44  & 1.83  & 15.13  & 84.02 &  77.71  & 1.83  & 14.43 \\

\cmidrule{1-10}

  \multirow{2}{*}{GRPO} & Qwen3-8B & 75.59 & 90.05 & 1.25 & 8.97 & 88.31 & 76.72 & 2.04 & 9.67 \\
  & Qwen3-14B & \underline{84.91} &  94.56 & \underline{1.87}  & 15.34 & \underline{90.27} & 78.50 & \underline{2.20} & 14.04  \\
  
\bottomrule
\end{tabular}
}
\caption{Performance comparison of various planning approaches across different LLMs. FPR and CPR are multiplied by 100. The \textbf{best} and \underline{second-best} results are highlighted. Time denotes the average inference time in seconds. \cellcolor{gray!25}{Values} with gray shading indicate inference times exceeding 40 seconds. 
}
\label{tab:planning-approaches}
\vspace{-0.5em}
\end{table*}

\section{Reinforcement Learning for Planning}
\label{grpo_process}

Travel plans live in a vast combinatorial space where supervised imitation generalizes poorly, as minor edits can flip an itinerary from feasible to invalid. We therefore turn to reinforcement learning. 
Beyond evaluation, the expert-validated TripScore reward can also serve as a dense, point-wise optimization signal that provides per-rollout supervision to discriminate between near-feasible variants. 

We adopt Grouped Relative Policy Optimization (GRPO)~\citep{DeepSeekMath} (Figure~\ref{fig:grpo}), which consumes fine-grained ordinal rewards directly and uses within-group normalization that is robust to the long-tailed reward distribution. For each query $q$ we sample $G$ rollouts $\{o_i\}_{i=1}^{G}$ from $\pi_{\theta_{\text{old}}}$, score each with our evaluator, and standardize within group: $\tilde{r}_i = (r_i - \mathrm{mean}(r))/\mathrm{std}(r)$. The token-level advantage is $\hat{A}_{i,t}{=}\tilde{r}_i$, and the policy is optimized with the clipped GRPO objective.

We score rollouts with the \emph{rule-only} component of TripScore. Putting LLM-based sub-scores inside the training loop would invite Goodhart-style hacking against a learnable judge, while rule-based scoring stays deterministic and verifiable. As shown in Section~\ref{sec:experiments}, gains from rule-only RL still transfer to the full evaluator and pairwise expert judgments, indicating that mastering verifiable feasibility is a sufficient foundation for downstream semantic quality.

\section{Experiments}
\label{sec:experiments}

\subsection{Experimental Setup}
We evaluate GPT-5.4, Claude Opus 4.6, Kimi-k2.5~\citep{kimi}, DeepSeek-V3.2~\citep{deepseek-r1}, and Qwen3-\{8B,14B,32B\}~\citep{Qwen3} across four method families: \emph{Direct}, \emph{Test-time compute} (ZS-CoT~\citep{zs-cot}, ReAct~\citep{ReAct}, LLM-Modulo~\citep{llm-modullo}, HyperTree~\citep{hypertree}), \emph{Neuro-Symbolic} (TTG~\citep{ttg}, NESY~\citep{ChinaTravel}), and \emph{Fine-tuning} (SFT, RFT~\citep{RFT}, GRPO~\citep{DeepSeekMath}). We report Format/Commonsense Pass Rate (FPR/CPR; CPR is conditioned on passing the format gate), Reward, and inference time (mixed closed-API and local inference, indicative only). Gemini-3-flash serves as the LLM judge for soft sub-metrics (temperature 0). All fine-tuning experiments (SFT, RFT, GRPO) report results from a single controlled run (seed 42); multi-seed variance is not characterized due to computational cost. FPR/CPR definitions and per-method implementation details are in Appendix~\ref{implementation}.

\subsection{Experiment Analysis}

\textbf{Test-time compute and solvers trade quality for latency.} In Table~\ref{tab:planning-approaches}, test-time methods lift feasibility but at prohibitive cost. For example, LLM-Modulo+GPT-5.4 achieves the top reward (3.13) and FPR (94.95\%) on TripScore-F yet quadruples inference time over Direct (106.7s vs.\ 27.70s), making it impractical for interactive deployment. Neuro-symbolic solvers (TTG, NESY) reach near-perfect commonsense (CPR $\approx$99\%) on the tiny format-passing subset but collapse on free-form FPR ($\sim$1.8\% for NESY on TripScore-F). The main reason is they treat ambiguous intent as hard constraints, leaving them over-constrained even with extensive prompt engineering and ample retry budgets (Appendix~\ref{neural_symbolic_failure}). We also find that test-time search is bounded by the base model's capability. HyperTree on Qwen3-8B drops reward from -1.00 (Direct) to -2.62, indicating that test-time compute amplifies the strength of the base model rather than substituting for it.

\textbf{RL fine-tuning internalizes planning.} GRPO offers the best trade-off between reward and latency. GRPO on Qwen3-8B reaches reward 2.04 on TripScore-F under 10s latency, surpassing Qwen3-8B Direct (-1.00) and even GPT-5.4 Direct (1.81). On Qwen3-14B, GRPO attains the second-best on TripScore-M and TripScore-F rewards while remaining over $7\times$ faster than LLM-Modulo+GPT-5.4. This supports internalizing planning behaviors as parametric knowledge being more compute-efficient than test-time search.

\subsection{Travel Plan Quality}

To disentangle quality from feasibility, we conduct a controlled pairwise comparison restricted to queries where both plans pass feasibility, pairing each Qwen3-14B variant against GPT-5.4. Each pair is scored both by the TripScore reward and by three experts (we keep only unanimous annotations). 

Table~\ref{tab:quality-pairwise} shows that SFT, RFT, and especially GRPO progressively narrow the quality gap with GPT-5.4. On TripScore-M under TripScore reward, GPT-5.4's win rate drops from 0.726 (vs.\ Direct) to 0.459 (vs.\ GRPO). These gains generalize well to free-form queries on TripScore-F (0.744 $\to$ 0.522). Human expert evaluation further corroborates this trend. Specifically, on TripScore-M the expert win rate narrows progressively: $0.701 \to 0.549 \to 0.504 \to 0.477$ (Direct $\to$ SFT $\to$ RFT $\to$ GRPO). For GRPO, the two-sided binomial test does not reject $H_0{:}\,p{=}0.5$ ($p{=}0.628$), though we note that failure to reject does not formally establish equivalence. Crucially, the reward and expert rankings move in lockstep across all four variants, providing evidence that TripScore tracks human-perceived quality.

\begin{table}[h]
\centering
\scriptsize
\setlength{\tabcolsep}{4pt}
\resizebox{0.45\textwidth}{!}{

\begin{tabular}{llcc}
\toprule
\textbf{Dataset} & \textbf{Opposing Method} & \textbf{Win Rate} & \textbf{$p$}  \\
\midrule
\multicolumn{4}{c}{\textbf{TripScore Reward}}           \\
\midrule
\multirow{4}{*}{TripScore-M}& Direct (Qwen3-14B)  & 0.726 & 0.002 \\
& SFT (Qwen3-14B)  & 0.587 & 0.632 \\
& RFT (Qwen3-14B)  & 0.515 & 1.000  \\
& GRPO (Qwen3-14B) & 0.459 & 0.484  \\
\midrule
\multirow{4}{*}{TripScore-F} & Direct (Qwen3-14B)  & 0.744 & 0.004 \\
& SFT (Qwen3-14B)  & 0.659 & 0.032 \\
& RFT (Qwen3-14B)  & 0.552 & 0.711  \\
& GRPO (Qwen3-14B) & 0.522 & 0.245  \\
\midrule
\multicolumn{4}{c}{\textbf{Expert Selection}}           \\
\midrule
\multirow{4}{*}{TripScore-M} & Direct (Qwen3-14B)  & 0.701 & 0.004 \\
& SFT (Qwen3-14B)  & 0.549 & 0.137 \\
& RFT (Qwen3-14B)  & 0.504 & 0.839  \\
& GRPO (Qwen3-14B) & 0.477 & 0.628  \\
\midrule
\multirow{4}{*}{TripScore-F} & Direct (Qwen3-14B)  & 0.785 & 0.002 \\
& SFT (Qwen3-14B)  & 0.614 & 0.008 \\
& RFT (Qwen3-14B)  & 0.561 & 0.636  \\
& GRPO (Qwen3-14B) & 0.513 & 0.492  \\
\bottomrule
\end{tabular}
}
\caption{Pairwise quality comparison: Direct (GPT-5.4) vs.\ Qwen3-14B variants. Win rate $>0.5$ favors GPT-5.4. The $p$ value is from a two-sided binomial test against $H_0{:}\,p{=}0.5$; $p>0.05$ indicates no statistically significant preference between the two methods.}

\label{tab:quality-pairwise}
\vspace{-0.5em}
\end{table}

\subsection{Evaluator Robustness Analysis}
We re-score outputs under Rule-only, LLM-only, and leave-one-out (LOO) ablations of individual sub-scores (full protocols and tables in Appendix~\ref{evaluator_ablation}). System-level Kendall's $\tau$ between Full TripScore and the ablated rankings remains high on TripScore-M (Rule-only $\tau{=}0.85$, LLM-only $\tau{=}0.60$), but on TripScore-F the Rule-only ranking flips to $\tau{=}{-}0.07$ while LLM-only stays strong ($\tau{=}0.88$). Removing Request Fulfillment alone collapses correlation to $\tau{=}0.01$, confirming that rigid rule compliance is insufficient for free-form queries. Our central claim remains stable across all variants. Specifically, GRPO (Qwen3-14B) matches GPT-5.4 on feasible plans (2.42 vs 2.43), outperforms it on Rule-only metrics (1.91 vs 1.88), and trails marginally on LLM-only semantic metrics (2.11 vs 2.15), indicating that RL fine-tuning bridges the gap primarily by mastering execution rules, providing a structural foundation for semantic quality.

\subsection{Production Deployment}
To validate real-world impact, we conducted an online A/B test comparing the GRPO-fine-tuned 14B model against the prior closed-source general LLM API that previously served all production traffic. Over 539K generated itineraries (425,784 from the prior API; 113,783 from GRPO), we measured plan adoptability (users accepting the generated plan) and constraint-aware adoptability (acceptance of plans satisfying stated constraints). GRPO achieves 16.1\% / 32.6\% on these metrics versus 7.35\% / 18.9\% for the prior API, a relative lift of +121\%. Based on these results, the GRPO model now fully replaces the closed-source endpoint in production.

\subsection{Can Code Agents Solve Planning?}
\label{sec:code_agent}

We test whether frontier code-execution agents can improve planning by writing scripts. On 50 random TripScore-M queries, Claude Code (v2.1.92, Opus 4.6) with an explicit tool-use prompt degrades FPR (84\% vs.\ 96\% direct), reward (1.27 vs.\ 2.48), and quintuples latency (204s vs.\ 38s; Table~\ref{tab:code_agent_main}). Inspecting all 50 traces, 84\% either skipped code or used tools trivially (e.g., \texttt{print}) despite explicit encouragement; only 7/50 wrote useful code (e.g., computing weekday), and 1/50 fell into an infinite loop. On a sole-planning task with all information in-context, scripting an ambiguous multi-objective routing problem is harder than reasoning natively---what limits performance is parametric planning competence, not tool-use proficiency.

\begin{table}[h]
\centering
\small
\setlength{\tabcolsep}{4pt}
\resizebox{0.35\textwidth}{!}{
\begin{tabular}{lcc}
\toprule
\textbf{Metric} & \textbf{Direct} & \textbf{Claude Code} \\
\midrule
FPR & 96.0\% & 84.0\% \\
CPR & 93.8\% & 73.8\% \\
Reward & 2.48 & 1.27 \\
Avg.\ Generation Time & 38.3s & 204.2s \\
\bottomrule
\end{tabular}
}
\caption{Direct prompting vs.\ Claude Code on 50 TripScore-M queries (same queries, same model). Full prompt and case analysis in Appendix~\ref{code_agent_case}.}
\label{tab:code_agent_main}
\end{table}

\section{Conclusion}

We introduce TripScore, a behavior-grounded benchmark and expert-validated reward calibrated against 1,468 expert pairwise judgments. Across prompting, test-time compute, neuro-symbolic solvers, code agents, and fine-tuning, RL on a 14B open-source model matches GPT-5.4 under expert evaluation, and is now deployed in production replacing the prior closed-source endpoint. 


\section*{Limitations}

We highlight several limitations of this work. First, our benchmark focuses on sole-planning given retrieved information, rather than end-to-end tool use. While this isolates planning competence, it may overestimate performance relative to interactive settings with retrieval errors, tool failures, and partial observability. Second, parts of TripScore rely on LLM-based sub-metrics (iconic landmark coverage, attraction diversity, free-form request fulfillment with Gemini-3-flash as judge). To mitigate Goodhart effects, our RL/RFT training uses the rule-only scorer, but the full evaluator may still inherit judge-specific biases (e.g., sensitivity to prompting, model version updates). Future iterations could ensemble multiple judges or ground semantic checks in more granular verifiable facts. Third, TripScore-F is smaller than TripScore-M (219 test queries), which may limit statistical power for fine-grained ablations and for detecting small effect sizes in pairwise comparisons. Expanding free-form coverage and auditing distribution shifts across locales, languages, and traveler profiles are important next steps. Finally, due to the high computational cost of fine-tuning and full-pipeline evaluation, we report results from a single run with a fixed random seed (42). Future versions should report variance across multiple seeds.



\bibliography{custom}

\appendix

\section{Web Application Usage}
\label{web_application}
\begin{figure*}[htbp]
    \centering
    \includegraphics[width=0.7\textwidth]{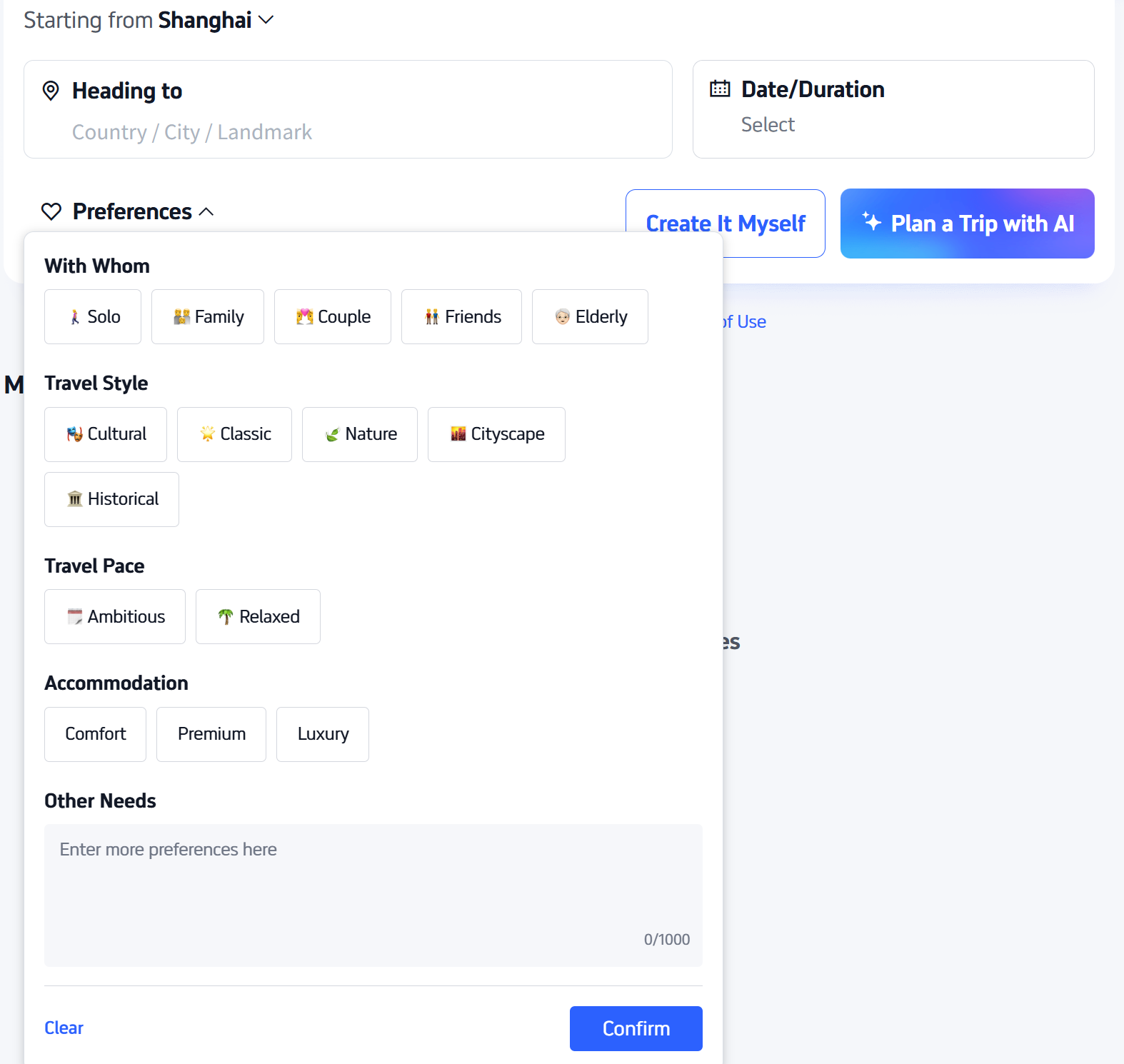}
    \caption{Main interface of the travel-planning web application used for collecting real-user interaction logs.
    }
    \label{fig:user_interface}
\end{figure*}

Our web application interface is designed to ensure effortless access for users worldwide. Figure~\ref{fig:user_interface} showcases the main screen through which users submit their travel requests and generate travel plans. Users can select destinations and duration, choose either enumerated preferences or type free text needs. Then users can click on ``Plan a Trip with AI'' to generate tailored travel plans.

To mitigate the risk of collecting low-quality data, we enforce several quality-control measures. First, users must register with an email or phone number for identity verification. Second, we audit logs in real time and take action to ban accounts that generate NSFW content, open multiple sessions or operate at abnormal speed. Third, we use an NSFW keyword list blocks to prevent users from harmful outputs. Last, each user is initially allowed at most 100 interactions per day.

\paragraph{Consent, privacy, and de-identification.}
Users are informed that their interactions may be used to improve the system and for research, and logging is enabled only after explicit consent. For research processing, we remove direct identifiers and store requests in a de-identified form. We do not include account identifiers, emails/phone numbers, or payment information in the benchmark data. We also exclude requests that contain obvious personally identifying content. The retained data---destinations, durations, general departure cities, and stated preferences---are common, non-unique attributes shared across large user populations and do not permit re-identification of individual users. Access to raw logs is restricted to authorized personnel, and the benchmark snapshot used in this paper is frozen (August 2025) to support versioned evaluation.

\section{Dataset Construction Pipeline}
\label{data_process}

\textbf{Real data from real users.} A key advantage of TripScore is that our data is collected from real, intrinsically motivated users, rather than paid crowd workers or AI. We developed a web application that allows users to initiate their own travel plans by selecting destinations and duration, and by providing either predefined preference constraints or free-form needs. With explicit user consent, we log their selections and requests, recording approximately ten thousand requests per day. In our logs analyzing over 300,000 user requests collected over a one-month period, 39.3\% of users select only destinations and duration, 38.3\% submit free-form customized requirements, and 22.4\% choose predefined preference options. Based on this distribution, we establish two distinct evaluation tracks. The first track strictly retains only destinations and duration, while the second track accommodates free-form user requests without predefined constraints.

\textbf{TripScore-M query selection.} To enhance authenticity, we incorporate real user requests from logs under appropriate permissions. We construct minimal-constraint queries by randomly sampling real user interactions where only destinations and trip duration are provided. While these users initially provide minimal input, implicit preferences often exist. We augment these queries with explicit preferences, selected from a predefined set outlined in the personal preference constraints in Table~\ref{tab:cons-description}, to test the planner's ability to handle such constraints when they are specified or inferred, ensuring the benchmark covers the full spectrum of planning complexity. 

\textbf{TripScore-F query selection.} We construct free-form queries from user interactions where they submit free-form text queries. These queries capture genuine planning intent rather than contrived prompts. These queries span a wide range, from specific tasks (e.g., attending a meeting), to day-by-day attraction preferences (e.g., Day 1: city, Day 2: zoo), and to theme-based trips (e.g., a Harry Potter trail). To maintain quality, we exclude requests under 10 words, as such inputs are overly simplistic and insufficiently challenging for models to generate tailored itineraries. Note that the TripScore-F set is kept relatively small (219 queries) due to the high cost of rigorous expert verification and privacy filtering.

\textbf{Relevant information recall.} To ensure adequate and accurate retrieval, we rely on production-grade industry data curated and maintained by professional operators. In deployment, an agent utilizes corresponding search tools to query the travel database (covering flights, hotels, and attractions) using keywords extracted from the user request. The retrieval process continues iteratively until relevant options for all requested categories are found or a maximum step limit is reached. The planning model then consumes both the collected information and the user request to produce the final plan. However, to streamline the evaluation process in this benchmark, we bypass the tool-call stage and directly provide the planning model with the information collected by the agent. This approach allows us to focus specifically on assessing the model's planning capabilities while ensuring access to all necessary data for generating final travel plans. 

\textbf{Quality control.} After collecting the user request and corresponding information, we implement a rigorous quality assurance process to remove the invalid request, leveraging both LLMs and human evaluation. We filter out requests with empty retrieved information, ensuring that each query has adequate information for planning. 
Then we use Gemini-3-flash to filter out samples whose relevant information is inconsistent with the user query.
After the automated filtering step, human annotators review the remaining pool to remove requests that are prohibitively difficult or nonsensical.

\section{Benchmark Details}
\label{data_details}

In Table~\ref{tab:bench-details}, we list the detailed data distribution on training, validation and test set.

\begin{table}[htbp]
    \centering


     

\resizebox{0.45\textwidth}{!}{
\begin{tabular}{lccccc}
    \toprule
    \textbf{Dataset} &
    \textbf{Type} &
    \textbf{\#Examples} &
    \textbf{\#City} &
    \textbf{\#Hotel} &
    \textbf{\#Attraction} 
    \\
    \midrule
     \multirow{1}{*}{Training} & TripScore-M & 3,493 & 781 & 7,830 & 9,788 \\
         \midrule

      \multirow{2}{*}{Validation} & TripScore-M & 100 & 70 & 1,057 & 1,525 \\
     & TripScore-F  & 58 & 41 & 638 & 1,313\\
          \midrule
          
       \multirow{2}{*}{Test} & TripScore-M & 1,000 & 358  & 4,459 & 5,452 \\
     & TripScore-F  & 219 & 122 & 1,894 & 2,057 \\
     \midrule

        \multicolumn{2}{l}{ \textbf{All}}  & 4,870 & 897 & 9,376 & 10,997 \\
    \bottomrule
\end{tabular}}
    \caption{Statistics of dataset distribution. Splits are random at the query level; no stratification by user, city, hotel, or attraction is applied. The City, Hotel, and Attraction columns report post-split descriptive statistics (entity counts), not splitting criteria.}
    \label{tab:bench-details}
\end{table}

\section{Constraints}
\label{constraints}
As illustrated in Table~\ref{tab:cons-description}, we evaluate plan feasibility and quality across four dimensions, which include format and commonsense constraints for feasibility checks, as well as soft and personal preference criteria for overall quality assessment.

Format constraints ensure the output is structurally parsable and grounded in the provided database. Commonsense constraints enforce physical and temporal validity, such as travel times and operating hours. Soft constraints capture general quality indicators like route efficiency and variety, while Personal Preference constraints measure alignment with specific user needs, ensuring the plan is not just valid, but desirable.

\begin{table}[htbp]
    \centering
    \resizebox{\linewidth}{!}{
\small
\begin{tabular}{lll}
\toprule
\textbf{Constraint}                         & \textbf{Description}       &       \textbf{Evaluation}                                                                                                                                                                \\ \midrule
\rowcolor[gray]{0.85}
\multicolumn{3}{c}{\textbf{\textit{Format Constraint}}}                                                                                                                                                                                 \\ \midrule
\addlinespace[0.25em]
Response Format                    & \parbox{0.85\linewidth}{All responses must follow the requested structure and organization exactly as specified.}    & Rule \\    

\addlinespace[0.25em]
Information Verification                & \parbox{0.85\linewidth}{All attractions, transportation, and hotels in the plan must come from the provided information, otherwise, it will be considered a hallucination.}                                                    & Rule                         \\

\addlinespace[0.25em]
Information Accuracy                    & \parbox{0.85\linewidth}{Details like name and departure time must match the provided information.}                                                                                                                             & Rule       \\ 

\addlinespace[0.25em]
\textcolor{blue}{Information Relevance}                    & \parbox{0.85\linewidth}{\textcolor{blue}{All descriptions must specifically match their intended attractions (e.g., not mixing up or blending details from different places).}}                  & \textcolor{blue}{Rule}     \\ 

\midrule
\rowcolor[gray]{0.85}
\multicolumn{3}{c}{\textbf{\textit{Commonsense Constraint}}}                                                                                                                                                                                 \\ \midrule

\addlinespace[0.25em]
Information Completeness          & \parbox{0.85\linewidth}{All necessary information must be included, especially accommodation for each destination with multi-day stay and essential transportation.}                  & Rule                                                                     \\

\addlinespace[0.25em]
Chronological Order  & \parbox{0.85\linewidth}{All activities must be listed in chronological order.}                                                                      & Rule       \\

\addlinespace[0.25em]
Location Consistency                & \parbox{0.85\linewidth}{Each day's activity must be scheduled in the city where the traveler is actually present and change only after any required transportation.}                           & Rule                                                                                              \\

\addlinespace[0.25em]
Operating Hours            & \parbox{0.85\linewidth}{Visits to attractions must only be scheduled during their confirmed opening hours.}          & Rule                                                                                                                \\

\addlinespace[0.25em]
\textcolor{blue}{Travel Block-Out}      & \parbox{0.85\linewidth}{\textcolor{blue}{No activities can be scheduled between departure and arrival times during transportation.}}  & \textcolor{blue}{Rule} \\


\addlinespace[0.25em]
Transport. Consistency    & \parbox{0.85\linewidth}{Arrange the transportation from the starting city to each destination in sequence to avoid jumps or repetitive routes.}  & Rule  \\

\midrule

\rowcolor[gray]{0.85}

\multicolumn{3}{c}{\textbf{\textit{Soft Constraint}}}                                                                                                                                   \\ \midrule

\addlinespace[0.25em]
\textcolor{blue}{Schedule Density}      & \parbox{0.85\linewidth}{\textcolor{blue}{Each day's plan should be thoughtfully paced, ensuring neither overly lengthy nor excessively brief periods of activity.}} & \textcolor{blue}{Rule} \\

\addlinespace[0.25em]
\textcolor{blue}{Hotel Consistency}  & \parbox{0.85\linewidth}{\textcolor{blue}{When staying in the same city, the same hotel should be used throughout to avoid unnecessary check-ins and check-outs.}}  & \textcolor{blue}{Rule} \\

\addlinespace[0.25em]
\textcolor{blue}{Daytime Utilization} & \parbox{0.85\linewidth}{\textcolor{blue}{Fill daytime hours when evening travel is planned; avoid starting the day with attractions that open late.}} & \textcolor{blue}{Rule} \\

\addlinespace[0.25em]
Unique Attractions & \parbox{0.85\linewidth}{Attractions should not appear more than once in the plan.} & Rule \\

\addlinespace[0.25em]
Location Clustering & \parbox{0.85\linewidth}{When possible, group attractions that are close to each other to reduce travel time and maximize sightseeing time.} & Rule \\

\addlinespace[0.25em]
\textcolor{blue}{Iconic Landmarks} & \parbox{0.85\linewidth}{\textcolor{blue}{Include renowned local attractions and must-see sites in the plan. } } & \textcolor{blue}{LLM} \\

\addlinespace[0.25em]
Attraction Diversity & \parbox{0.85\linewidth}{Avoid overrepresentation of similar attractions to ensure a varied and engaging itinerary.} & LLM \\

\midrule

\rowcolor[gray]{0.85}

\multicolumn{3}{c}{\textbf{\textit{Personal Preference Constraint}}}                                                                                                                                                                                        \\ \midrule

\addlinespace[0.25em]
Budget Preference                         & \parbox{0.85\linewidth}{The plan should align with the user’s budget expectations (eg. ``premium'', ``budget-conscious'', or ``best value'').}    & Rule                                                                                  \\
\addlinespace[0.25em]
\textcolor{blue}{Pacing Preference}                        & \parbox{0.85\linewidth}{\textcolor{blue}{The plan should reflect the user's desired pacing (eg. ``relaxed'', ``moderate'', or ``compact''). }} & \textcolor{blue}{Rule}                                                                                       \\
\addlinespace[0.25em]
\textcolor{blue}{Attraction Prioritization}                          & \parbox{0.85\linewidth}{\textcolor{blue}{The plan should prioritize or include the user-specified types of attractions.} }               & \textcolor{blue}{Rule}                                                                 \\ 

\addlinespace[0.25em]
\textcolor{blue}{Physical Effort Preference}                        & \parbox{0.85\linewidth}{\textcolor{blue}{The plan should balance walking distances and physical intensity, matching the user’s indicated effort level (eg. ``light'', ``moderate'', or ``strenuous'').} }       & \textcolor{blue}{Rule}                              \\ 





\addlinespace[0.25em]
User Request Fulfillment             &    \parbox{0.85\linewidth}{The plan should follow the user's specific request.}    & LLM      \\ 

\bottomrule

\end{tabular}

}

    \caption{Constraint description. The format constraints assess whether the plan adheres to the specified format guidelines. The commonsense constraints evaluate the logical feasibility of the plan. The soft and preference constraints measure the alignment of the plan with plan quality standards and user preferences. Constraints in black originate from existing benchmarks, whereas constraints in \textcolor{blue}{blue} are our newly proposed constraints. Exact formulas and parsing rules for each metric are detailed in Appendix~\ref{soft_rule} and Appendix~\ref{preference_rule}.}
    \label{tab:cons-description}
\end{table}

\section{Constraints Comparison}
\label{constraints_comparison}
Constraints across benchmarks are summarized in Table~\ref{tab:cons-comparison}, showing which dimensions are evaluated and by what mechanism.
TripScore distinguishes itself by covering a broader spectrum of constraints compared to prior benchmarks like TravelPlanner or ChinaTravel. While previous works primarily focus on hard constraints (feasibility), TripScore integrates a comprehensive set of soft and preference constraints to evaluate plan quality holistically. Furthermore, our hybrid evaluation mechanism uniquely combines deterministic rule-based checks for objective criteria with LLM-based assessments for semantic and subjective qualities, providing a more robust and expert-aligned evaluation signal.

\begin{table*}[htbp]
    \centering
    \resizebox{\linewidth}{!}{
\small

\begin{tabular}{llcccc}
\toprule

\textbf{Dimension} & \textbf{Constraints} & \textbf{Ours (Eval)} & \textbf{ChinaTravel (Eval)} & \textbf{TravelPlanner (Eval)} & \textbf{TripTailor (Eval)} \\
\midrule
\multirow{4}{*}{Format}
& Response format  & \greencheck  (Rule) & \greencheck  (Rule) & \greencheck  (Rule) & \greencheck  (Rule) \\
& Info verification  & \greencheck  (Rule) & \greencheck  (Rule) & \greencheck  (Rule) & \greencheck  (Rule) \\
& Info accuracy & \greencheck  (Rule) & \greencheck  (Rule) & \greencheck  (Rule) & \greencheck  (Rule) \\
& Info relevance & \greencheck  (Rule) &  \redcross &  \redcross &  \redcross \\
\midrule
\multirow{6}{*}{Commonsense}
& Info completeness & \greencheck  (Rule) & \greencheck  (Rule) & \greencheck  (Rule) & \greencheck  (Rule) \\
& Chronological order & \greencheck  (Rule) & \greencheck  (Rule) &  \redcross &  \redcross \\
& Location consistency & \greencheck  (Rule) & \greencheck  (Rule) & \greencheck  (Rule) & \greencheck  (Rule) \\
& Operating hours & \greencheck  (Rule) & \greencheck  (Rule) &  \redcross &  \redcross \\
& Travel block-out  & \greencheck  (Rule) &  \redcross &  \redcross &  \redcross \\
& Transport consistency  & \greencheck  (Rule) & \greencheck  (Rule) & \greencheck  (Rule) & \greencheck  (Rule) \\
\midrule
\multirow{7}{*}{Soft}
& Schedule density  & \greencheck  (Rule) &  \redcross &  \redcross &  \redcross \\
& Hotel consistency  & \greencheck  (Rule) &  \redcross &  \redcross &  \redcross \\
& Daytime utilization & \greencheck  (Rule) &  \redcross &  \redcross &  \redcross \\
& Unique attractions & \greencheck  (Rule) & \greenhalfcheck (Rule) &  \redcross & \greenhalfcheck (Rule) \\
& Location clustering  & \greencheck  (Rule) & \greenhalfcheck (Rule) &  \redcross & \greenhalfcheck (Rule) \\
& Iconic landmarks coverage & \greencheck  (LLM) &  \redcross &  \redcross &  \redcross \\
& Attraction diversity& \greencheck  (LLM) &  \redcross &  \redcross &  \redcross \\
\midrule
\multirow{5}{*}{Preference}
& Budget style  & \greencheck  (Rule) & \greenhalfcheck (Rule) & \greenhalfcheck (Rule) & \greenhalfcheck (Rule) \\
& Pacing style & \greencheck  (Rule) &  \redcross &  \redcross &  \redcross \\
& Attraction prioritization& \greencheck  (Rule) &  \redcross &  \redcross &  \redcross \\
& Physical effort & \greencheck  (Rule) &  \redcross &  \redcross &  \redcross \\
& User request & \greencheck  (LLM) &  \redcross &  \redcross & \greencheck  (LLM) \\
\bottomrule
\end{tabular}

}
    \caption{Dimension-wise coverage of evaluation constraints across benchmarks. Cells indicate whether a constraint is evaluated and by which mechanism (Eval: Rule or LLM). \greencheck\ = evaluated; \greenhalfcheck\ = similar constraint exists; \redcross\ = not evaluated.}
    \label{tab:cons-comparison}
\end{table*}

\section{Expert Validation}
\label{exper_score}

\subsection{Expert Annotation}
\label{exper_annotation}

We conducted a comparative evaluation, integrating expert assessments with our automated framework. We utilized DeepSeek-V3.2, Qwen3-32B and GPT-5.4 to independently generate travel plans based on 3,000 user queries. After filtering out plans that failed to satisfy format and commonsense constraints, we obtained a final total of 1,468 pairwise comparisons. This final dataset consists of 267 comparisons between Qwen3-32B and DeepSeek-V3.2, 303 between Qwen3-32B and GPT-5.4, and 898 between GPT-5.4 and DeepSeek-V3.2.

Then we tasked 203 travel specialists with ranking these pairs of travel plans and providing rationales for their choices. For each pair, the specialists evaluated three options, namely favoring route A, favoring route B, or indicating that neither route met satisfactory standards. The specialists are verified professionals who earn their living by designing customized travel plans for clients and possess domain knowledge of specific destinations. We verified their credentials through employment verification with partner travel agencies and required a minimum of 2 years of professional planning experience. All annotators were compensated at professional rates via partner-agency contracts, commensurate with their regular consulting compensation. Each specialist was assigned plans featuring destinations with which they were familiar. Each plan pair was independently evaluated by three expert annotators.

After annotation, we computed inter-annotator agreement using Cohen's $\kappa$ for pairwise comparisons and Fleiss' $\kappa$ for multi-rater reliability. The results demonstrated moderate agreement among the experts, with Cohen's $\kappa$ reaching 0.5421 for pairwise comparisons and Fleiss' $\kappa$ achieving 0.5039 for multi-rater reliability. 
Furthermore, we evaluated inter-rater reliability. The results demonstrate an average pairwise agreement of 71.69\% between annotators, while the overall consensus across all three raters stands at 59.00\%.

\subsection{Grid Search}
\label{weight_optimize}
We employ a grid search optimization to find the optimal weight for our evaluation framework. The search space consists of 13 parameters: 7 soft constraint weights ($w_1$), 4 preference weights ($w_2$), and 2 multiplier weights ($w_3$, $w_4$). We discretize the parameter space using coarse grids: $w_1 \in \{0.1, 0.2, 0.4, 0.5, 0.7\}$, $w_2 \in \{0.2, 0.6, 1.0\}$, $w_3 \in \{0.8, 1.0, 1.2\}$, and $w_4 \in \{0.1, 0.4, 0.6, 0.8, 1.0, 1.2, 1.4\}$. The optimization is performed jointly for TripScore-M and TripScore-F tracks, and we set track-specific parameters for different tracks, shown in Table~\ref{tab:optimized_weights}. The optimized weight values are provide in Table~\ref{tab:optimized_weights}. While some optimal weights lie on the search space boundaries (e.g., $w_4=0.1$ for TripScore-M), we found that extending the range (e.g., to 0.05) did not yield statistically significant improvements in expert agreement on the validation set, suggesting the current boundaries are sufficient to capture the relative importance of these factors without overfitting.

We conduct nested stratified 5-fold cross-validation with an inner-loop grid search, yielding validation accuracy $0.6175 \pm 0.0259$. A single held-out split gives accuracy $0.6029$ with $\tau{=}0.2071$. On the full 1,468-pair set, 1{,}000$\times$ bootstrap gives accuracy $0.6031$ with a 95\% CI of $[0.5343,\,0.6667]$ and Kendall's $\tau{=}0.2073$ with a 95\% CI of $[0.0712,\,0.3377]$.

\subsection{Sensitive Analysis}
\label{sensitive_analysis}

Moreover, we probe robustness by jointly scaling the penalty multipliers $(w_3,w_4)\in\{(0.5,0.05),(1.0,0.10),(1.5,0.15),(2.0,0.20)\}$ and by applying simplex normalization (including temperature smoothing) to the soft and preference constraint weights. Across all variants, train agreement remains in the range 59.40\%-62.26\% and validation agreement in 58.47\%-61.32\%, i.e., within $\approx\!2.8$ pp (train) and $\approx\!2.9$ pp (val) absolute of the baseline. The near-constant validation accuracy under multiplier scaling and simplex reparameterizations indicates the method is robust to these design choices and does not rely on a narrow hyperparameter configuration.

\begin{table}[h]
\centering

\resizebox{0.5\textwidth}{!}{
\begin{tabular}{llcc}
\toprule
\multicolumn{2}{l}{\textbf{Parameter}} & \textbf{TripScore-M} & \textbf{TripScore-F} \\
\midrule
\multirow{7}{*}{\textbf{w1} (soft constraints)}  & Schedule Density & 0.70 & 0.70 \\
 & Hotel Consistency & 0.50 & 0.50 \\
 & Daytime Utilization & 0.40 & 0.40 \\
 & Unique Attraction & 0.20 & 0.20 \\
 & Location Clustering & 0.70 & 0.70 \\
 & Iconic Landmark & 0.10 & 0.10 \\
 & Attraction Diversity & 0.20 & 0.20 \\
\midrule
\multirow{4}{*}{\textbf{w2} (preferences)}
 & Attraction Prioritization & 0.20 & - \\
 & Pacing & 0.60 & - \\
 & Budget & 0.60 & - \\
 & Physical Effort & 0.60 & - \\
 & User Request & - & 1.00 \\
    \midrule
\textbf{w3} (soft constraint multiplier) & & 1.00 &  1.00 \\
\textbf{w4} (preference multiplier) & & 0.10 &  1.40 \\
\bottomrule
\end{tabular}}
\caption{Optimized weights value. Due to the different evaluation method, minimal-constraint (TripScore-M) and free-form (TripScore-F) datasets share the weights for soft constraints but utilize distinct preference weights.}

\label{tab:optimized_weights}
\end{table}

\subsection{Noise-adjusted Analysis}
\label{noise_adjusted}

Under a standard K-class symmetric-noise model, we can relate the average human-human pairwise agreement $A_{\text{pair}}$ and a typical annotator's agreement with respect to the latent truth $r$ as follows:

\begin{equation}
A_{\text{pair}} = r^2 + \frac{(1-r)^2}{K-1}
\end{equation}

\begin{equation}
r = \frac{1 + \sqrt{(K-1)(K \cdot A_{\text{pair}}-1)}}{K}
\end{equation}

Where $K$ is the number of classes. With $K = 3$ and $A_{\text{pair}} = 0.7169$, we calculate the human reliability ceiling, $r \approx 0.8390$. Then the expected model-human agreement $A_{\text{model},h}$ relates to the model's latent-truth agreement $r_{\text{model}}$ by:

\begin{equation}
A_{\text{model},h} = r_{\text{model}} \cdot r + \frac{(1-r_{\text{model}})(1-r)}{K-1}
\end{equation}

Which can be rearranged to solve for $r_{\text{model}}$:

\begin{equation}
r_{\text{model}} = \frac{ K \cdot A_{\text{model},h} - A_{\text{model},h} + r - 1}{ K \cdot r - 1}
\end{equation}

Substituting $A_{\text{model},h} = 0.6029$, $K=3$ and $r \approx 0.8390$ yields $r_{\text{model}} \approx 0.689$. Thus, the raw 60.29\% agreement corresponds to a noise-adjusted latent agreement of approximately 68.9\%. We can express this as a ratio of the model's performance to the human reliability ceiling $ r_{\text{model}} / r \approx 0.821 $. This indicates that the model achieves about \textbf{82.1\%} of the human reliability ceiling.

As a verification, we can also calculate the predicted three-annotator all-agree rate $A_{\text{all}}$ by the calculated human reliability ceiling $r$:

\begin{equation}
A_{\text{all}} = r^3 + \frac{(1-r)^3}{(K-1)^2} \approx 0.594
\end{equation}

This closely matches the observed overall agreement of 0.590 between all annotators, providing a sanity check for our calculations.

\subsection{Evaluation Comparison Experiments}
\label{evaluation_accuracy}

We implement two LLM-as-judge paradigms to evaluate pairs of candidate travel plans and evaluate agreement with expert labels as well as rank correlation (Kendall's $\tau$). Both paradigms use a structured rubric of hard/soft constraints and explicitly condition on the user request. To improve reproducibility, we use frozen prompts, temperature 0, and report results across 3 runs. And we report the results for multiple LLMs (GPT-5.4, DeepSeek-V3.2, Gemini-3-flash/pro).

\textbf{Point-wise scoring.}
The LLM independently scores each plan in $[0,100]$ under the same request and rubric. To mitigate instability reported in prior work~\citep{llm-judge2}, we adopt comparative prompting, anchoring scores with comparative references. The predicted winner is the plan with the higher score, and ties yield a neither decision. We report two point-wise metrics, namely tie-excluded accuracy and tie-inclusive accuracy. The full prompt is provided in Appendix~\ref{point_wise}.

\textbf{Pair-wise comparison.} The LLM receives the request and both travel plans, first eliminates candidates with hard-constraint violations, then compares soft quality and preference matching. If still uncertain, it selects the clearer, more executable plan. The model must output exactly one token from {route A, route B}. The prompt is provided in Appendix~\ref{pair_wise}.

\textbf{Ours scoring method.} For our hybrid evaluation framework combining rules and LLMs, we apply the identical configuration used in point-wise scoring. Specifically, the plan with the higher score is designated as the winner, while tied scores result in no decision.
We employ various LLMs for the LLM-based component. Because tie-excluded and tie-inclusive accuracies are nearly identical, we report only the tie-inclusive accuracy in Table~\ref{tab:evaluation_experiment}.

\section{Experiment Details}

\begin{figure*}[htbp]
    \centering
    \includegraphics[width=0.9\textwidth]{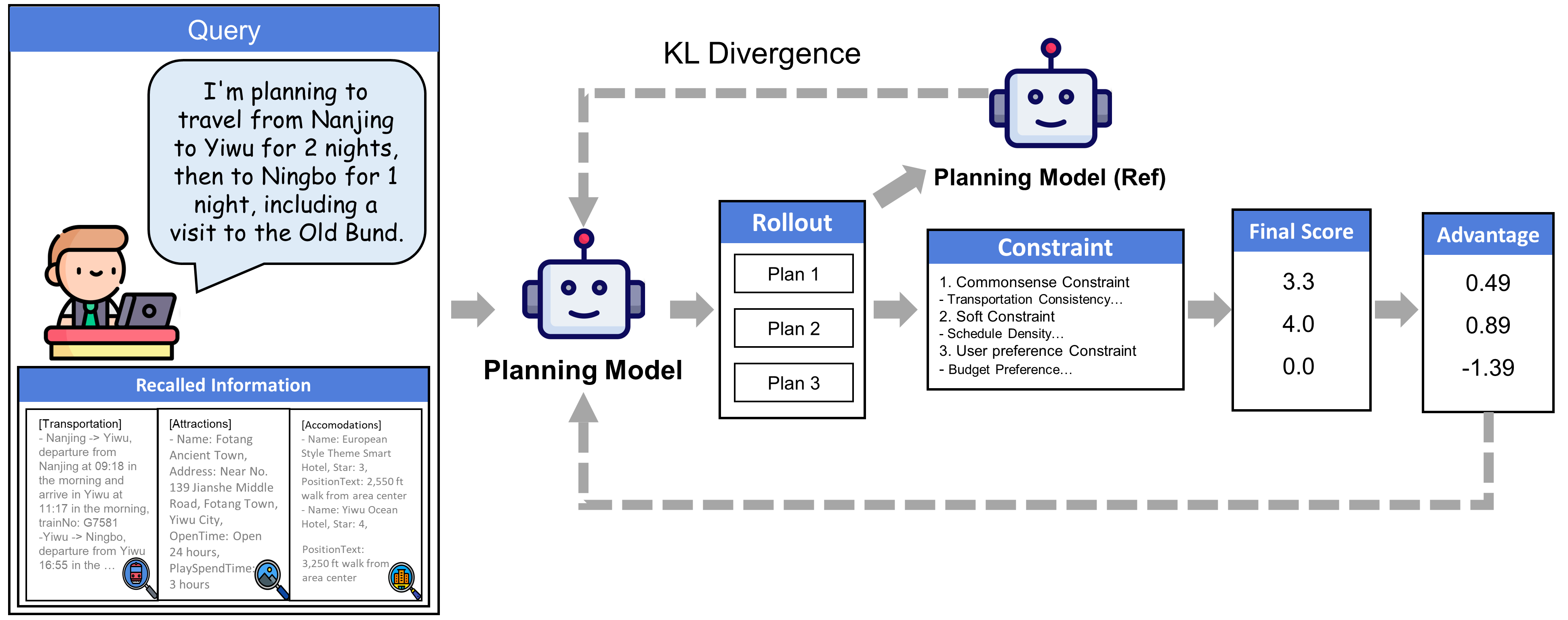}
    \caption{Demonstration of the Grouped Relative Policy Optimization (GRPO). GRPO estimates the baseline from group scores, and normalize them into a standard value. The planning model undergoes iterative optimization by maximizing the objective function $\mathcal{L}(\pi_\theta)$.
    }
    \label{fig:grpo}
\end{figure*}

\label{implementation}

\subsection{Fine tuning}

\textbf{SFT.} 
We fine-tune the pretrained model with pairwise query-response supervision. The SFT training set is constructed by prompting GPT-5.4 to answer queries from the TripScore-M training split. Then we filter out samples that violate our format or commonsense constraints. This yields 2,094 training and 71 validation examples. As GPT-5.4 does not expose an explicit chain-of-thought, we train the Qwen3 base model in a non-thinking configuration. During training, we set the maximum sequence length to 115,000 tokens and training epochs to 3. As the context length is too long, which may lead to GPU out of memory, we set the sequence parallel to 8 to ensure training progress. 



\textbf{RFT~\citep{RFT}.}  
We further refine the SFT-tuned Qwen3 models via RFT. Starting from the queries in the TripScore-M training split, we prompt the SFT model to generate five candidate routes per query and score each route with our own evaluator. The highest-scoring route is retained; queries for which all five candidates fail the quality bar are dropped. This curation leaves 2,372 clean training samples. We then fine-tune the SFT model for three epochs on this set, using the same hyperparameters as SFT.


\textbf{GRPO~\citep{DeepSeekMath}.} 
We train the SFT model based on GRPO algorithm on queries from the TripScore-M training set. Given that the SFT model was trained without a thinking mode, we preserve this setting and continue fine-tuning it using the GRPO algorithm. For each query, the policy model generates 8 rollouts, and rewards are computed by our evaluation framework. To speed up training, we use only the rule-based components of our evaluator. During training process, the maximum prompt length is 79000 and the maximum answer length is 7000.



We perform early stopping by selecting the best-performing checkpoint for each task independently. The hyperparameters employed in fine-tuning baselines are presented in Table~\ref{tab:train-details}.

\begin{table}[ht]
    \centering

     

    
          

\resizebox{0.5\textwidth}{!}{
\begin{tabular}{lccc}
    \toprule
    \textbf{Method} &
    \textbf{Hyperparameter} &
    \textbf{Qwen3-8B}  &
    \textbf{Qwen3-14B}  
    \\
    \midrule
     \multirow{5}{*}{SFT} &  learning rate & 2e-5 & 1e-5 \\
      & scheduler type &  cosine &  cosine \\
       & batch size &  32 &  32 \\
     & training epoch & 3 & 3  \\
       & warmup ratio &  0.1  & 0.1 \\
     
     \midrule

  \multirow{5}{*}{RFT} &   learning rate & 2e-5 & 1e-5 \\
    &  scheduler type &  cosine &  cosine \\
    &  batch size &  32 &  32 \\
    & training epoch & 3  & 3  \\
    &   warmup ratio &  0.1 &  0.1 \\
    
          \midrule
          
     \multirow{6}{*}{GRPO} &  actor learning rate & 1e-6  & 1e-6 \\
    &  scheduler type &  constant &  constant \\
    &  batch size &  64 & 64 \\
    &  training epoch & 1 & 1  \\
    & rollout temperature &  1.2 &  1.2 \\
    & rollout times & 8  & 8 \\
    \bottomrule
\end{tabular}
}

    \caption{The hyperparameters we employ in baselines.}
    \label{tab:train-details}
\end{table}

For training, we use the veRL~\citep{verl} framework and fine-tune the base models in 32 A100 GPUs. For deployment, we leverage vLLM~\citep{vllm} to serve Qwen3-8B, Qwen3-14B and Qwen3-32B. To maximize inference speed, we set tensor parallelism to 8 and run each model on a single 8 A100 node. All training experiments were conducted using a fixed random seed of 42 to ensure reproducibility. Due to the high computational cost of fine-tuning and testing, we report results from a single run.

\subsection{Other Baselines}
\textbf{Direct.} This approach involves inputting the query directly into the model, accompanied by comprehensive instructions detailing the task requirements and all relevant gathered information. The model is expected to generate a response based solely on this input. For Qwen3 base models, we disable the thinking mode for the direct method.

\textbf{Zero-Shot Chain-of-Thought (ZS-CoT)~\citep{zs-cot}.} This method enhances the reasoning process by encouraging the model to articulate intermediate steps. Building upon the Direct method, we augment the prompt with the phrase ``Let's think step by step.'' This addition is designed to elicit a more detailed, structured reasoning process from the model, potentially leading to more accurate and transparent outcomes. For Qwen3 base models, we enable the thinking mode for the ZS-COT method.

\textbf{ReAct~\citep{ReAct}.} This method incorporates environmental feedback into the planning process. We provide the function FeasibleEnquiry to check day-level feasibility to help enhance the model's reasoning ability. We develop specialized prompts to guide the LLM in generating daily travel plans and refining them based on the function feedback. Once the model calls the Finish function, the latest plan is returned as the final output. To control the time cost, LLMs are allowed to call the function up to 5 times. Otherwise, it is regarded as failing to generate the plan.

\textbf{LLM-Modulo~\citep{llm-modullo}.} 
LLM-Modulo is a hybrid method where a large language model (LLM) generates structured representations from a user's request, and a symbolic planner verifies them and flags issues in a feedback loop. The method benefits from the bidirectional interaction: The symbolic module actively gives feedback to the LLM, and the LLM refines the itinerary iteratively.

To implement the LLM-Modulo into our benchmark, we develop specialized prompts to guide the LLM in generating initial itineraries and refining them based on evaluator feedback. We reuse our existing constraint evaluators to automatically score itineraries and provide structured feedback to the LLM, and for fairness we only utilize the rule-based part in our evaluation framework. We also add a controller that coordinates iterations between the LLM and the evaluators until constraints are satisfied or a timeout is reached. We set a maximum of 3 iteration steps and stop the refinement process if the reward score exceeds 3.5.

\textbf{HyperTree~\citep{hypertree}.} This baseline employs a hypertree-structured reasoning paradigm that recursively decomposes a travel query into transportation, accommodation, and attraction subtasks. By employing this recursive decomposition strategy, the baseline ensures that each aspect of the travel plan can be meticulously crafted and seamlessly integrated, resulting in a highly personalized and adaptable itinerary.

To adapt HyperTree to our dataset, we implement several modifications. We remove all the hypertree rules and nodes related to Dining to align with our schema. We also refactor the hypertree library to handle an arbitrary-length destination list, dynamically generating city-specific subtrees and their corresponding inter-city transportation segments, with no upper bound on the number of cities. Additionally, as the repository does not provide prompts, we write them from scratch based on the descriptions in the paper. 

\textbf{TTG~\citep{ttg}.} 
TTG (To The Globe) is a hybrid travel-planning system that translates natural language requests into structured JSON constraints using a fine-tuned LLM. It then solves these constraints with a Mixed Integer Linear Programming (MILP) solver to guarantee feasible and near-optimal itineraries in under 5 seconds. It combines the LLMs' natural language abilities with the mathematical guarantees of MILP solvers. 

To implement TTG into our benchmark, we define a symbolic type to represent feasible travel constraints, serving as the intermediate format for MILP problem generation. Then we translate travel requests into MILP formulations that encode itinerary feasibility and optimization objectives. Finally, we integrate the Python \texttt{PuLP} library to solve the constructed MILP problems using appropriate solver strategies.

\textbf{NESY~\citep{ChinaTravel}.} The NESY baseline scheme consists of a two-stage process: (1) In the NL2DSL translation stage, natural language queries are converted into logical and preference DSL requirements; (2) In the interactive search stage, a neuro-symbolic solver sequentially arranges activities under the guidance of a symbolic sketch and LLM-driven POI recommendations, generating a multi-day itinerary with DSL validation. 

To adapt NESY to our dataset, we implement the following modifications:

\begin{itemize}[label={$\bullet$}, leftmargin=1em]
\item \textbf{Dataset Adaptation}. We streamline the planning process by removing redundant logical modules irrelevant to our study, such as restaurants, ensuring the workflow aligns with the available data support.

\item \textbf{DSL and Concept Function Reconstruction}. We reconstruct the concept functions and augmented the DSL statements with commonsense constraints to match the specific format of our itinerary results. Specifically, we removed DSL statements related to restaurants, people number, dishes, and room types, while expanding commonsense constraints. Ultimately, 32 concept functions were obtained, and the input of each function includes the planned itinerary elements  as well as relevant information.

\item \textbf{Multi-Destination Planning}. To address the limitations of the original NESY framework, which only supported single-destination planning and lacked mechanisms for sequencing multiple destinations, we implement a two-phase modification:
(1) For \textbf{determining the destination sequence}, we prioritize transportation data (e.g. direct accessibility, travel duration) to determine the planning order. In the absence of such data, we default to the sequence extracted from user input via LLMs.
(2) For \textbf{day allocation}, we introduce a ``cyclic allocation method'': each destination is initially assigned one day, with additional days distributed sequentially according to the predetermined order until meeting the total day requirement. 

\item \textbf{Transportation Mode Adaptation}. To address the original method's limitation of deeming planning infeasible when large-scale transportation data (e.g. flights, trains) are unavailable, we implement a more flexible approach. In the absence of such data, we now default to a ``self-driving'' mode. This adaptation accommodates scenarios like local trips and self-driving tours, maximizing the utilization of POI resources in areas lacking comprehensive transportation data. We establish fixed time windows for self-driving based on typical travel patterns: outbound trips are scheduled from 9:00 to 11:00, and return trips from 18:00 to 20:00.
\end{itemize}

\section{Case Study}

\begin{figure*}[h!]
    \centering
    \includegraphics[width=0.85\textwidth]{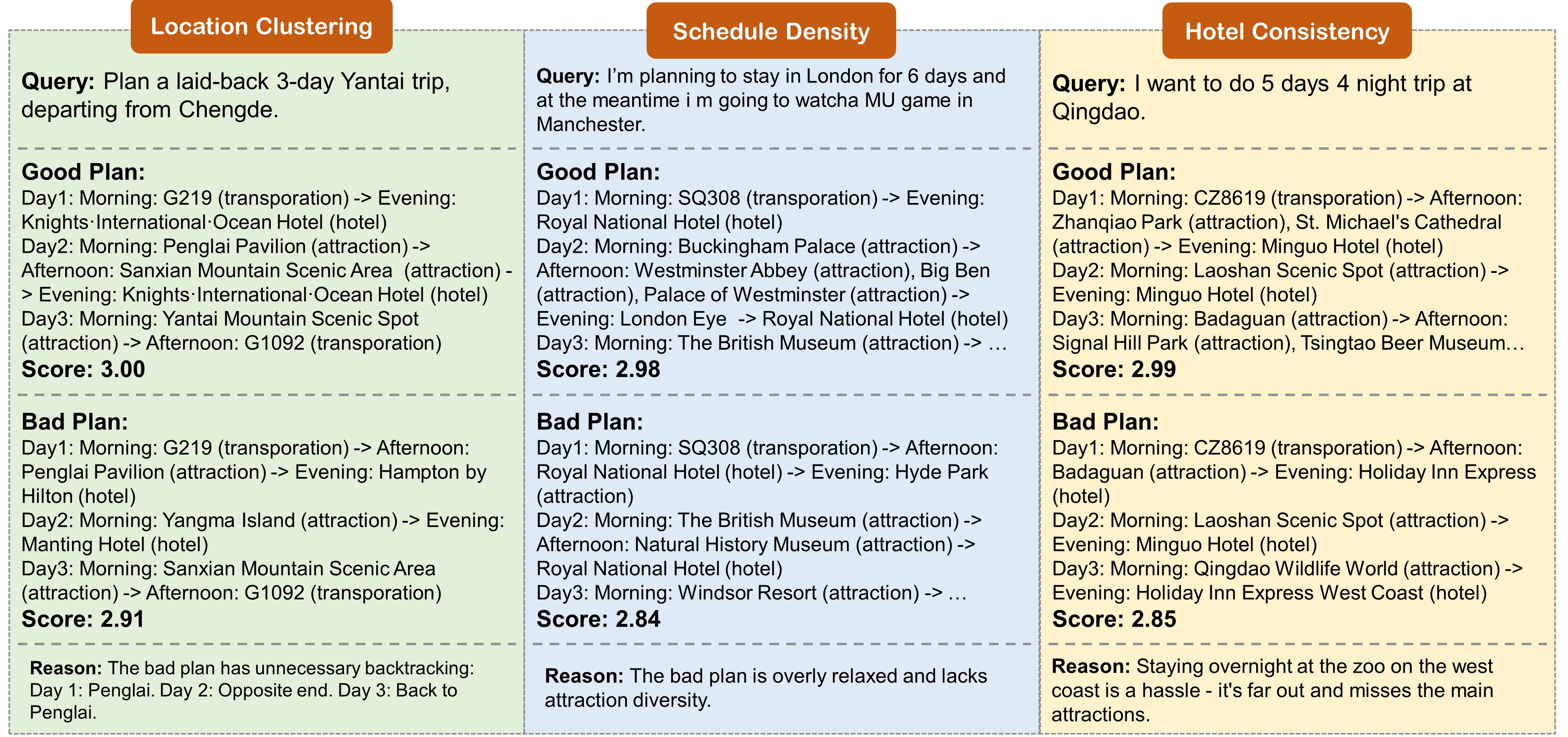}
    \caption{Case studies of the comparison of the plan quality. 
    }
    \label{fig:casestudy}
\end{figure*}

To examine how plan quality is assessed, we present several head-to-head cases in Figure~\ref{fig:casestudy}. Although all plans are feasible, side-by-side comparison under our criteria clearly exposes which route is superior. This underscores the need for fine-grained comparative evaluation rather than relying solely on rules or single-pass LLM judgments. Reliable evaluation must account for structural validity, spatio-temporal coherence, semantic value (iconicity and diversity), and user preferences. A practical evaluator should fuse rule-based diagnostics with preference signals, quantify uncertainty, and return actionable feedback.

\section{Evaluator Ablations and Robustness Protocol}
\label{evaluator_ablation}

This section presents the results of the robustness protocol described above. We re-scored the outputs of all models from Table~\ref{tab:planning-approaches} using the evaluator variants and analyzed the stability of rankings and conclusions.

To quantify the agreement between different evaluator variants, we compute the system-level Kendall's $\tau$ correlation. To ensure a fair comparison of plan quality independent of feasibility failures, we adopt a pairwise intersection protocol. For each planning method, we compare it against a reference baseline (Direct GPT-5.4) and identify the subset of queries where both methods produced feasible plans (passing format and commonsense checks). We then compute the average score of the target method on this common subset under each evaluator variant. The Kendall's $\tau$ coefficient is calculated between the system ranking induced by the Full TripScore and the ranking induced by the ablated variant, measuring how well the ablated metric preserves the relative order of model performance on valid plans.

\subsection{Ranking Stability Results}

\begin{table}[h]
\centering

\small
\resizebox{0.49\textwidth}{!}{
\begin{tabular}{lcc}
\toprule
\textbf{Evaluator Variant} & \textbf{TripScore-M} (\(\tau\)) & \textbf{TripScore-F} (\(\tau\)) \\
\midrule
\textbf{A1. Component Ablation} & & \\
Rule-only (Feasibility + Rule Soft) & 0.85 & -0.07 \\
LLM-only (Feasibility + LLM Soft) & 0.60 & 0.88 \\
\midrule
\textbf{A2. Leave-One-Out (LOO)} & & \\
LOO-Schedule Density & 0.85 & 0.96 \\
LOO-Hotel Consistency & 0.95 & 0.96 \\
LOO-Daytime Utilization & 0.96 & 0.96 \\
LOO-Unique Attractions & 0.97 & 0.97 \\
LOO-Location Clustering & 0.95 & 0.96 \\
LOO-Iconic Landmarks & 0.93 & 0.99 \\
LOO-Attraction Diversity & 0.91 & 0.84 \\
LOO-Structured Preference & 0.47 & 1.00 \\
LOO-Request Fulfillment & 1.00 & 0.01 \\
\bottomrule
\end{tabular}
}
\caption{System-level ranking correlations (Kendall's $\tau$) between Full TripScore and ablated variants.}
\label{tab:ablation_kendall}
\end{table}

Table~\ref{tab:ablation_kendall} highlights the necessity of hybrid evaluation. In the minimal-constraint setting (TripScore-M), both rule-based and LLM-based components correlate moderately with the full score (0.85 and 0.60), and removing structured preferences drops correlation to 0.47, indicating a balanced contribution of feasibility rules and preference matching.


\subsection{Claim Stability: GRPO vs. GPT-5.4}
We further investigated the nature of the improvements brought by RL fine-tuning. While Table~\ref{tab:quality-pairwise} compares GRPO against Direct (Qwen3-14B), a more ambitious goal is to bridge the gap with proprietary SOTA models. We compared the conditional quality scores (on feasible plans in TripScore-F) of GRPO (Qwen3-14B) against Direct (GPT-5.4) across evaluator variants.

\begin{table}[h]
\centering

\small
\resizebox{0.49\textwidth}{!}{
\begin{tabular}{lccc}
\toprule
\textbf{Evaluator Variant} & \textbf{GRPO (Qwen3-14B)} & \textbf{Direct (GPT-5.4)} & \textbf{Gap} \\
\midrule
\textbf{Full TripScore} & 2.42 & 2.43 & \(\approx 0.00\) \\
\textbf{Rule-only} (Feasibility + Rule Soft) & \textbf{1.91} & 1.88 & \textcolor{green}{+0.03} \\
\textbf{LLM-only} (Feasibility + LLM Soft) & 2.11 & \textbf{2.15} & -0.04 \\
\bottomrule
\end{tabular}
}
\caption{Average conditional scores on TripScore-F (Generalized) for feasible plans. GRPO (14B) matches GPT-5.4 on the full score, effectively closes the gap by surpassing GPT-5.4 on rule-based execution while remaining competitive on semantic understanding.}
\label{tab:ablation_scores}
\end{table}

As shown in Table~\ref{tab:ablation_scores}, GRPO achieves parity with GPT-5.4 on the Full TripScore (2.42 vs 2.43). The decomposition reveals the mechanism: GRPO \textbf{outperforms GPT-5.4 on Rule-only metrics} (1.91 vs 1.88), indicating superior adherence to strict constraints (format, operating hours, schedule density). On \textbf{LLM-only metrics} (iconic, diversity, request fulfillment), GRPO trails slightly (2.11 vs 2.15) but remains highly competitive. This confirms that RL fine-tuning effectively internalizes complex planning rules, enabling a 14B model to rival GPT-5.4's quality on real-world queries, driven primarily by robust constraint satisfaction.

\section{Neuro-Symbolic Solver Case Study}
\label{neural_symbolic_failure}
\lstset{
    basicstyle=\footnotesize\ttfamily,
    breaklines=true,
    breakatwhitespace=true,
    breakindent=0pt,
    columns=fullflexible,
    keepspaces=true,
    showstringspaces=false,
    xleftmargin=0pt,
    aboveskip=4pt,
    belowskip=4pt,
    commentstyle=\color{gray!80!black}\itshape,
    keywordstyle=\bfseries,
}

The high failure rate of NeSy-based approaches stems from a fundamental mismatch between rigid symbolic logic and the ambiguous, flexible nature of real-world travel planning. While NeSy systems excel at verifiable logic, they often lack the semantic flexibility to handle ``soft" constraints, translating vague user intents into overly strict, unexecutable logical predicates. We illustrate this with three specific failure modes observed in our experiments:
 
\subsection{Semantic Misinterpretation}
Users frequently request activities that are not discrete physical locations (POIs) in a database, such as ``attend a meeting" or ``rest." NeSy translators, lacking semantic grounding, often erroneously map these abstract concepts to rigid POI-visit constraints.

Example Query: ``Trip around Beijing. I have a meeting all day tomorrow and want to rest on Thursday''

Generated DSL Constraint:
\begin{lstlisting}[language=Python]
# The solver interprets "Meeting" and "Rest" as specific POI names to be visited
result_day2 = (day2_attraction_names_set <= {'Meeting'})
result_day3 = (day3_attraction_names_set <= {'Rest'})
\end{lstlisting}
Failure Cause: The solver attempts to find attractions literally named ``Meeting" or ``Rest" within the candidate set. Since no such entities exist and the solver cannot produce entities outside the candidate set, it returns ``Unsatisfiable." In contrast, an end-to-end LLM would correctly interpret this as a directive to leave a time slot empty or instantiate a custom event outside the provided candidate set.

\subsection{Rigid Entity Matching}
Symbolic solvers typically enforce strict set membership or equality checks. If a user provides a generic term (e.g., ``Jiuzhaigou hotels") while the database uses specific official names, the rigid constraint fails to bridge the semantic gap.

Example Query: ``Jiuzhaigou 3-day trip. Jiuzhaigou hotels and attractions.''

Generated DSL Constraint:
\begin{lstlisting}[language=Python]
# The solver requires the selected hotel's name to be strictly "Jiuzhaigou"
result = ({'Jiuzhaigou'} <= hotel_names_set)
\end{lstlisting}
Failure Cause: Real-world database entries are specific (e.g., ``InterContinental Resort Jiuzhaigou Paradise"). The above strict logical condition evaluates to False. The solver lacks the flexibility to perform semantic fuzzy matching or understand that the user implies ``any hotel located in Jiuzhaigou."

\subsection{Oversimplified Logic Generation}
NeSy translators often struggle to convert complex, implicit constraints (like ``overnight train") into accurate logical predicates, resorting to brittle heuristics that fail to capture the true condition.

Example Query: ``Trip starting from Guangzhou, March 27th-31st. I want to spend the night on the train.''

Generated DSL Constraint:
\begin{lstlisting}[language=Python]
# The solver simplifies "overnight on train" to a naive time check
overnight_train_found = False
if departure_time >= '23:00' or arrival_time <= '06:00':         
    overnight_train_found = True
result = overnight_train_found
\end{lstlisting}
Failure Cause: The generated logic is logically flawed or too restrictive. It naively assumes ``overnight" means departing after 11 PM or arriving before 6 AM. It misses standard overnight trains that might depart at 8 PM and arrive at 8 AM the next day. Consequently, valid overnight options are filtered out, leading to unnecessary plan failures.

These examples highlight that hard constraints demand a level of precision in prompt-to-logic translation that is inherently brittle in open-domain scenarios. NeSy solvers are blocked by their own logical strictness when facing ambiguity, whereas LLMs can employ soft reasoning to resolve these gaps.

\section{Code Agent Case Study}
\label{code_agent_case}
This appendix provides the full system prompt used for the Claude Code experiment in Section~\ref{sec:code_agent}. To give the agent harness its best chance, we explicitly modified the system prompt to encourage tool usage:

\begin{quote}\small
\texttt{[Tool Usage] You have access to a code execution environment. For complex planning, constraint checking, budget calculation, or data processing, you may choose to write and execute Python code to assist you. However, using code is optional. You may also reason and generate the itinerary directly. Regardless of the method you choose, your final output must strictly be the requested JSON.}
\end{quote}

All other settings (the planning task instructions, the input context, and the JSON output schema) are identical to the Direct setting, ensuring that the only difference between the two conditions is the agent harness itself.

\section{Error Distribution Breakdown}
\label{error_dist}

\begin{figure}[h!]
    \centering
    \includegraphics[width=0.5\textwidth]{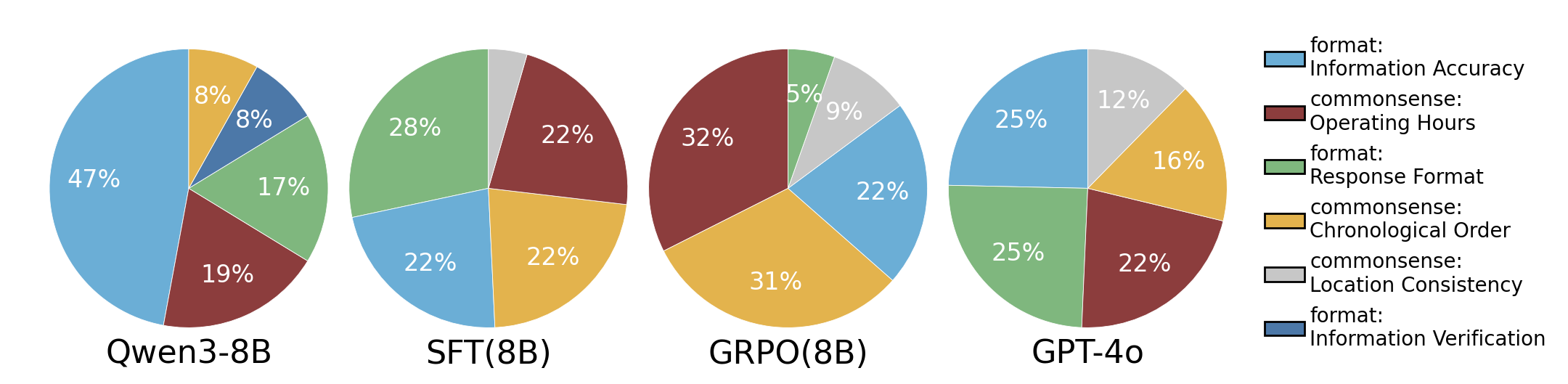}
    \caption{Format and commonsense constraints error distribution on the TripScore-F test set.
    }
    \label{fig:error_analysis}
\end{figure}

To analyze the pass rate, Figure~\ref{fig:error_analysis} breaks down violations of format and commonsense constraints on the TripScore-F set. For clarity, we report the top 5 most frequently broken constraints for each method. 
Furthermore, we found that the consistently dominant errors across all methods stem from format issues related to information accuracy and response structure, along with commonsense violations regarding operating hours and chronological order.
Notably, Qwen3-8B exhibits the highest proportion of information-accuracy failures. These typically manifest as hallucinations, such as mismatched attraction IDs and names. Both SFT and GRPO training methodologies demonstrate significant efficacy in mitigating this type of hallucinations.

\section{Performance on Various Trip Durations}

\begin{figure}[h]
    \centering
    \includegraphics[width=0.48\textwidth]{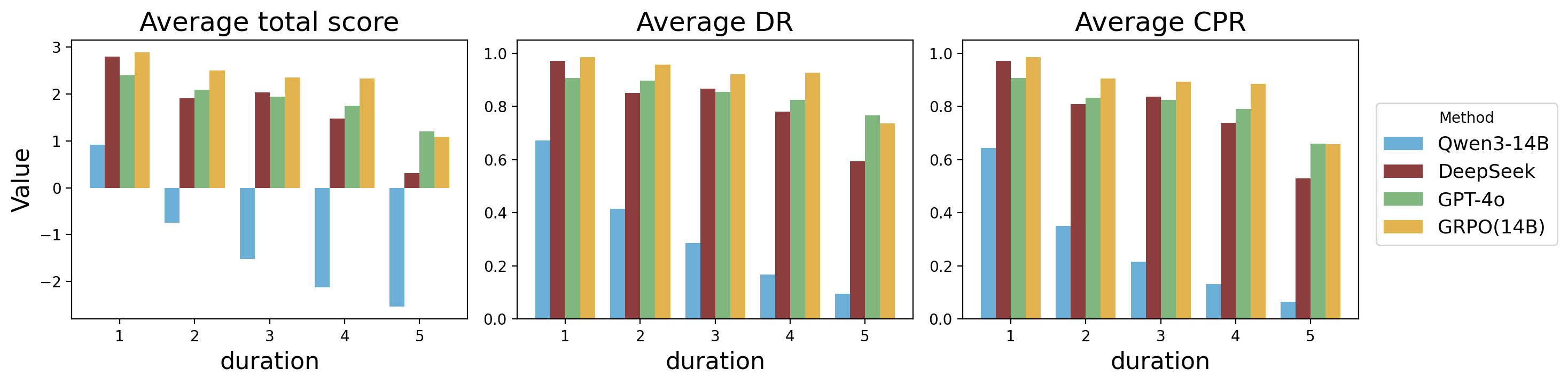}
    \caption{Performance distribution across varying trip durations on the TripScore-M test set.}
    \label{fig:complexity}
    \vspace{-0.5em}
\end{figure}

Figure~\ref{fig:complexity} reports model performance as task complexity increases (longer trip durations). To maintain statistical significance, we focus on results for trips ranging from 1 to 5 days, excluding longer durations with insufficient data points. As the trip duration extends from 1 to 5 days, we observe a notable degradation in performance across vanilla baselines. This decline is evident in both format constraint (Average FPR) and commonsense constraint (Average CPR), with a corresponding decrease in reward. In contrast, GRPO (Qwen3-14B) generally achieves the highest FPR and CPR scores while exhibiting the smallest performance decline as trip duration increases, indicating superior stability in long-horizon planning scenarios. This suggests that GRPO learns to maintain consistency and constraint satisfaction while preserving valid output structure, leading to stronger performance when scheduling becomes more challenging. 


\section{Soft Constraint Score Rule}
\label{soft_rule}




The soft-constraint vector $\bm{S}_{\text{soft}}=(S_i)_{i=1}^{7}$ comprises seven sub-scores defined below: schedule density ($S_{\text{schedule}}$), hotel consistency ($S_{\text{hotel}}$), daytime utilization ($S_{\text{daytime}}$), unique attractions ($S_{\text{unique}}$), location clustering ($S_{\text{location}}$), iconic landmark ($S_{\text{iconic}}$), and attraction diversity ($S_{\text{diversity}}$).

\textbf{1. Schedule Density.} This metric assesses the temporal feasibility of daily itineraries by checking minimum and maximum activity hours against day-specific thresholds. Scoring is by day-level violation (at most one penalty per day):

\begin{equation}
S_{\text{schedule}}=1-\frac{ |d_v| }{D}
\end{equation}
where $D$ is the total number of days. $d_v$ is the set of days where the total activity hours exceed the upper bound (10 hours) or fall below the lower bound (4 hours).

\textbf{2. Hotel Consistency.} The hotel consistency metric applies penalties to unnecessary hotel changes within a single city:

\begin{equation}
S_{\text{hotel}} = 1 - \frac{|S_{\text{switches}}|}{|H|}
\end{equation}

where $S_{\text{switches}}$ represents the number of nights involving a switch to a nearby hotel (within 100 km) in the same city, and $|H|$ is the total number of hotel nights.

\textbf{3. Daytime Utilization.} This constraint ensures the efficient utilization of daytime hours, with a focus on avoiding idle time or overcrowded scheduling:

\begin{equation}
S_{\text{daytime}} = 1 - \frac{\sum_{d=1}^{D} \mathbf{1}[\text{violation}_d]}{D}
\end{equation}

where $\text{violation}_d$ occurs when there is not arranged any activities in the morning and afternoon.

\textbf{4. Unique Attractions.} The uniqueness score employs a combination of linear and exponential penalties to quantify the degree of redundancy among selected attractions:

\begin{equation}
\resizebox{0.85\hsize}{!}{$
\begin{aligned}
S_{\text{unique}} = \max \left(0, 1 - \frac{|A_{\text{dup}}|}{|A_{\text{total}}|}  - \sum_{a \in A_{\text{dup}}} \frac{(n_a - 1)^2 \cdot 0.05}{|A_{\text{total}}|} \right)
\end{aligned}
$}
\end{equation}

where $A_{\text{dup}}$ represents non-consecutively duplicated attractions. $n_a$ is the number of times an activity appears. $A_{\text{total}}$ represents the total attractions in the itinerary.

\textbf{5. Location Clustering.} This metric optimizes the spatial distribution of daily activities by applying penalties proportional to inter-attraction distances:

\begin{equation}
S_{\text{location}} = \max\left(0, 1 - \frac{\mathcal{P}_{\text{total}}}{|A_{\text{total}}|}\right) \\
\end{equation}

where $\mathcal{P}_{\text{total}}$ represents the count of consecutive activity pairs for the same day that are among the top 20\% farthest of all activity pairs in the itinerary. $|A_{\text{total}}|$ is the total number of activities.

\textbf{6. Iconic Landmark} and \textbf{Attraction Diversity}. These two metrics are assessed using a 5-point Likert scale, with evaluations conducted by LLMs. The resulting scores are normalized to the range $[0,1]$ so that they are comparable to other soft sub-scores. The corresponding prompts are provided in Appendix~\ref{diversity_prompt} and Appendix~\ref{iconic_prompt}. The normalization applied to both metrics is:

\begin{equation}
S = \frac{\text{rating} - 1}{4}, \quad \text{rating} \in \{1, 2, 3, 4, 5\}.
\end{equation}

This mapping ensures that the lowest rating (1) corresponds to a normalized score of 0, while the highest rating (5) yields a normalized score of 1.

\section{Preference Constraint Score Rule}
\label{preference_rule}

\textbf{TripScore-M Datasets.} The preference evaluator measures how well an itinerary aligns with the user's stated preferences. We consider four dimensions: $\bm{S_{\text{pref}}} = (S_{\text{budget}}, S_{\text{pacing}}, S_{\text{attraction}}, S_{\text{effort}})$.

1. Budget Preference ($S_{budget}$):
The budget preference metric calculates the proportion of consistency between users' budget-corresponding expected hotel star ratings and actual hotel star ratings, aiming to measure the matching degree of budget and hotel star:

\begin{equation}
S_{\text{budget}}=\frac{1}{|H|}\sum_{h\in H} g(h;\,\text{pref}).
\end{equation}

where $H$ is the set of hotel nights, and $g(h;\,\text{pref}) \in\{0,1\}$ indicate whether night $h$ matches the budget preference (e.g., 0-2 stars for "cost-effective", 3-4 stars for "comfortable" and 5 stars for "High-end").

2. Pacing Preference ($S_{\text{pacing}}$):
Pacing Preference quantifies the consistency between the actual itinerary arrangement and users' pace preferences based on key indicators such as play duration and the number of activities:

\begin{equation}
S_{\text{pacing}}=\frac{1}{D}\sum_{d=1}^D f(d;\,\text{pref}).
\end{equation}
where $D$ is the number of days and $f(d;\,\text{pref})\in[0,1]$ is a day-level compliance function. Specifically, $f(d;\,\text{pref}) = 1$ if the daily activity time is within the preferred range (e.g., 6-8 hours for "Moderate"), $f(d;\,\text{pref}) = 0.5$ if it deviates by less than 2 hours, and $0$ otherwise.

3. Attraction Preference ($S_{\text{attraction}}$):
We map user attraction preferences to POI tags and measure coverage among visited POIs. It is calculated by the proportion of qualified attraction:

\begin{equation}
S_{\text{attraction}}=\frac{1}{|A_{\text{total}}|}\sum_{a\in A_{\text{total}}} m(a;\,\text{pref}).
\end{equation}

where $A_{\text{total}}$ represents all visited attractions and $m(a;\,\text{pref}) \in\{0,1\}$ indicating whether attraction $a$ matches any preferred tag.

4. Physical Effort Preference ($S_{\text{effort}}$):
Calculates activity's physical effort value by rules, determines itinerary exertion type via high-effort activity and duration ratios, and quantifies consistency with users' physical effort preferences. 

\begin{equation}
S_{\text{effort}}=\frac{1}{D}\sum_{d=1}^D h(d;\,\text{pref}).
\end{equation}

where $ h(d;\,\text{pref}) \in\{0,1\}$ indicate whether the day $d$ matches the preferred physical effort tag. A day is labeled strenuous if (i) its single-day physical-exertion score is greater than 2, where hiking, theme-park and mountain-climbing each contribute 1, cycling contributes 2, and all other activities contribute 0; or (ii) it requires physical exertion on two consecutive days. A day is light if it involves no exertion and neither adjacent day involves exertion. All remaining cases are moderate.



\textbf{TripScore-F Datasets.} For real-world datasets, we employ LLM-based evaluation to assess overall user request compliance. $S_{pref} = \{S_{user}\}$. The corresponding prompt is provided in Appendix~\ref{user_prompt}.

\begin{equation}
\resizebox{0.85\hsize}{!}{$
\begin{aligned}
S_{user} = \alpha \text{LLM}(\text{prompt}, \text{response}, \text{user\_request})
\end{aligned}
$}
\end{equation}

where $\alpha$ represents the scaling weight (0.2) employed to normalize the score within the range of 0 to 1.

\section{Prompt List}
\label{prompt_list}
\subsection{Iconic Landmarks Evaluation Prompt}
\label{iconic_prompt}

\begin{tcolorbox}
[title= ICONIC\_LANDMARKS\_EVALUATION\_PROMPT,colback=blue!10,colframe=blue!50!black,arc=1mm,boxrule=1pt,left=1mm,right=1mm,top=1mm,bottom=1mm, breakable]
\small

Please evaluate whether the attractions in the following itinerary cover the classic must-visit attractions of corresponding destination.

\vspace{1em}

Itinerary:
\{answer\_text\}
\vspace{1em}

Please evaluate based on the following criteria:

1. Does it include the most famous landmark attractions of the destination.

2. Does it cover different types of classic attractions (historical culture, natural scenery, modern architecture, etc.).

3. The popularity and recommendation level of the attractions.

4. Please consider the number of days in the itinerary. If some secondary attractions cannot be covered due to insufficient days, you can relax the evaluation criteria.
\vspace{1em}

Please return the evaluation result in JSON format:

\{

\ \ \ \ "score": score (integer rating 1-5, where 1=no classic attractions, 2=only a few classic attractions, 3=some classic attractions, 4=most classic attractions, 5=all classic attractions),

\ \ \ \ "missing\_attractions": ["list of missing important classic attractions"],

\ \ \ \ "explanation": "detailed explanation"

\}

\end{tcolorbox}

\subsection{Attraction Diversity Evaluation Prompt}
\label{diversity_prompt}

\begin{tcolorbox}
[title= ATTRACTION\_DIVERSITY\_EVALUATION\_PROMPT,colback=blue!10,colframe=blue!50!black,arc=1mm,boxrule=1pt,left=1mm,right=1mm,top=1mm,bottom=1mm, breakable]
\small
Please evaluate the richness and diversity of the following itinerary to determine if there are homogenization issues.
\vspace{1em}

Itinerary:
\{answer\_text\}
\vspace{1em}

Please evaluate based on the following criteria:

1. Diversity of attraction types (historical culture, natural scenery, entertainment, shopping \& dining, etc.).

2. Richness of activity experiences (sightseeing, hands-on experiences, interactive, leisure, etc.).

3. Reasonable pace arrangement (balance of active/quiet, indoor/outdoor).

4. Avoiding repetitive or homogeneous activities.

5. Please consider the main attraction types of the destination. If the main attractions of the destination are of a single type, you can relax the evaluation criteria for homogeneity issues.
\vspace{1em}

Please return the evaluation result in JSON format:

\{

\ \ \ \ "score": score (integer rating 1-5, where 1=homogenization problem accounts for more than 80\% of the itinerary, 2=homogenization problem accounts for about 60\% of the itinerary, 3=homogenization problem accounts for about 40\% of the itinerary, 4=homogenization problem accounts for about 20\% of the itinerary, 5=homogenization problem is small or nonexistent),

\ \ \ \ "diversity\_issues": ["list of identified homogenization or monotony issues"],

\ \ \ \ "explanation": "detailed explanation"

\}

\end{tcolorbox}

\subsection{User Request Fulfillment Evaluation Prompt}
\label{user_prompt}

\begin{tcolorbox}
[title= USER\_PREFERENCE\_CONSTRAINT\_PROMPT,colback=blue!10,colframe=blue!50!black,arc=1mm,boxrule=1pt,left=1mm,right=1mm,top=1mm,bottom=1mm, breakable]
\small
You are a professional travel itinerary evaluation expert, responsible for evaluating whether the generated itinerary meets the user's specific request and expectations.

Please evaluate the following travel itinerary based on the assessment criteria to determine whether it meets the user's specific request and expectations.

\vspace{1em}

**User Request: **

\{user\_request\}

\vspace{1em}

**Generated Itinerary Response: **

\{answer\_text\}

\vspace{1em}

You need to carefully analyze the user's requirements and evaluate the itinerary's alignment based on the following aspects: Departure/Destination, Schedule/Timing, Mode of Transportation, Number of Travelers, Accommodation Requirements, Coverage of Attractions, Activity Types, Pace of the Trip, Budget, Other Requirements.

\vspace{1em}

**Scoring Criteria**

5 points: Excellent. The itinerary fully meets all the user's requirements and considers potential personalized needs, providing a travel plan that exceeds expectations.

4 points: Good. The itinerary fully meets all the user's core requirements; however, there are details that could be further optimized.

3 points: Average. The itinerary satisfies most user requirements, such as mandatory budget, schedule, and number of travelers, but some aspects are not adequately addressed.

2 points: Poor. The itinerary fails to meet the user's main requirements, with most elements misaligned with their preferences.

1 point: Very Poor. The itinerary completely fails to meet the user's expectations and is irrelevant to their request.

0 points: The user did not provide any specific information (e.g., "Plan a trip for me"), in which case any itinerary offered can be considered as meeting the user's needs.

\vspace{1em}

**Instructions for Scoring**

1. Your evaluation should focus on determining whether the provided itinerary meets the user's expectations.

2. If IDs are provided for transportation, POIs, or hotels, you may assume these details are authentic and reliable.

3. Before assigning a score, analyze the itinerary and the user’s request, explaining why you assigned that score.

4. If the user's request changes midway, base your evaluation on the latest requirements.

5. You only need to evaluate the current itinerary. If the user requests multiple or alternative options, this should not result in a deduction.

6. Strictly follow the JSON format below when providing the evaluation result

Output format:

\{

\ \ \ \ "detailed\_feedback": "Detailed evaluation feedback",
 
\ \ \ \ "final\_score": Final score (0-5)
    
\}

\end{tcolorbox}

\subsection{Itinerary Generation Prompt}

\begin{tcolorbox}
[title= ITINERARY\_GENERATION\_PROMPT,colback=blue!10,colframe=blue!50!black,arc=1mm,boxrule=1pt,left=1mm,right=1mm,top=1mm,bottom=1mm, breakable]
\small
You are a travel planning expert, skilled at generating detailed travel itineraries based on user's needs and preferences, and ultimately outputting them in JSON format.
Attention: User's requirements may change, you need to adapt to the latest query of the user.
\vspace{1em}

[Itinerary Arrangement Rules]

1. Arrange the itinerary according to the user's requirements. Don not change the user's travel plan, including the number of days, travel cities, travel dates and etc.

2. The itinerary should be arranged in chronological order, and the time period should be divided into Morning/Afternoon/Evening.

3. A well-designed itinerary should include transportation arrangements, accommodation and key attractions, all organized in a proper chronological sequence to ensure a smooth travel experience, Pay special attention to the restrictions on the opening hours of attractions (openTimeCalendar field) to avoid scheduling visits during times when the attractions are closed or not allowing entry. 

4. For transportation, prioritize choosing main stations and pay special attention to the timing of transportation arrangements, ensuring they are scheduled within intended time period. Avoid mismatches such as planning morning departures for afternoon or evening time period.

5. For accommodation, find the most suitable hotel from the most suitable hotel area. One city only need one hotel and do not change hotel in the same city.

6. The hotel should be arranged at the end of the day, and hotel arrangements should be indicated every night, except on the last day.

7. It is especially important to pay attention to the time requirements for attractions and transportation to avoid time conflicts, which could lead to an unreasonable or unachievable itinerary.

8. Ensure that the travel schedule and physical exertion are moderate, and avoid arranging too many activities in the same time period. 

9. Transportation for the outbound and return trips is required, and be careful not to mix up the outbound and return trips.

10. If the user's travel duration is short (less than 2 days), it is recommended to focus on visiting the core attractions within the city.

\vspace{1em}

[Reference Data Rules]

Provided reference data may include:

    Attractions: poi (id: poiId, name: poiName)
    
    Hotels: hotel (id: hotelid, name: hotelname)
    
    Transportations: train/flight/bus/driving/ferry/ship (id: planid, name: trainNo/flightNo/shipName)
    
you must select hotels and transportation from reference data. But attractions are allowed to use external resources if more suitable.

\vspace{1em}

[Json Format Instruction]

1. Extract exact name, type, and id when using reference data:

    For attractions type: `poi', name: poiName, id: poiId. 
    
    For hotels type: `hotel', name: hotelname, id: hotelid; 
    
    For transportations type: `transportation', name: trainNo/flightNo, id: planid.

2. Attractions which is not in the reference data are allowed, but must set ``id'': ``'' and when mentioning the attraction in the description, use the format **attraction**. transportation and hotel must be chosen from the reference data.

3. for the items has the same id, just output one item, do not repeat items with different name but same id.

4. the Json format is as follows:

\{

\ \ ``itineraryName'': ``itinerary name like: 3 Days's Travel Itinerary: Shanghai to Beijing'',

\ \ ``recommendReason'': ``the reason why this itinerary is recommended, and make user feel that this itinerary is very suitable for him/her requirements. recommend reason should be no more than 50 words'',
 
\ \ ``dayInfos'': 

\ \  \  [

 \ \ \ \    \{
    
 \ \ \ \   \ \   ``day'': ``the order of days,a integer number starting from 1'',
        
\ \ \ \   \ \   ``scheduleTitle'': ``today's schedule title'',
        
\ \ \ \    \ \    ``scheduleDetail'': 
        
\ \ \ \     \ \         [
                        
\ \ \ \ \ \    \ \   \{ 
                            
\ \ \ \ \ \   \ \        ``period'': ``the time period when the schedule begins, must choose one from Morning/Afternoon/Evening(Capitalized Initial Letter)'',

\ \ \ \ \ \     ``description'':  ``Mention all attractions/hotels/transportations using the specified markdown syntax: For attractions and hotels,you should point out names and ids, use the format   \verb|**[PoiName](poiId)**| or \verb|**[HotelName](hotelId)**|, if the attractions is not in the reference data, use the format \verb|**[PoiName]**|.''
                                               
\ \ \ \ \ \     \ \    \ \    \ \   ``detailList'':  
                                
\ \ \ \ \ \     \ \      \ \  \ \       [
                                                
\ \ \ \ \ \  \ \        \ \    \ \  \ \    \ \      \{
                                                    
\ \ \ \ \ \  \ \         \ \       \ \  \ \    \ \      \ \      ``type'': ``transportation/poi/hotel'',
                                                        
\ \ \ \ \ \  \ \      \ \     \ \  \ \    \ \     \ \            ``id'': ``planid/poiId/hotelid'',
                                                        
\ \ \ \ \ \  \ \        \ \     \ \  \ \    \ \    \ \      ``name'': ``trainNo/flightNo/poiName/hotelname''
                                                        
\ \ \ \ \ \  \ \     \ \      \ \    \ \    \ \         \}
                                                    
\ \ \ \ \ \           \ \      \ \    \ \        ]
                                                
\ \ \ \ \ \       \     \      \}
                            
\ \ \ \      \ \    ]
                        
\ \   \ \    \}
    
\ ],

\  ``tips'': 

\ \  \{

\ \ \ \   ``title'': ``tips title'',

\ \ \ \    ``info'': ``the tips’s total content should be within 50 words''

\ \    \}
    
\}

\vspace{1em}
[References]

[transportation arrangements]

\{transportation\_information\}
\vspace{1em}

[attractions reference information]

\{attraction\_information\}
\vspace{1em}

[hotels reference information]

\{hotel\_information\}

\vspace{1em}

[User Query]

\{user\_request\}

\end{tcolorbox}

\subsection{ReAct Prompt}

\begin{tcolorbox}
[title= REACT\_PROMPT,colback=blue!10,colframe=blue!50!black,arc=1mm,boxrule=1pt,left=1mm,right=1mm,top=1mm,bottom=1mm, breakable]
\small

You are a travel planning expert, skilled at producing a detailed itinerary based on the given transportation scheme and the user's requirements and preferences.

Please complete the task by alternating between Thought, Action, and Observation steps.

The Thought phase is used to reason about the current situation. The Action phase has two types:

(1) FeasibleEnquiry [Sub Plan]: used to assess whether a sub-plan is basically feasible. You must input the sub-plan in JSON format. The sub-plan should cover a full day.

(2) Finish [Final Plan]: used to indicate task completion. You must submit the complete and final plan (in JSON format) as the argument.

\vspace{1em}

[Itinerary Arrangement Rules]

1. Arrange the itinerary according to the user's requirements. Don not change the user's travel plan, including the number of days, travel cities, travel dates and etc.

2. The itinerary should be arranged in chronological order, and the time period should be divided into Morning/Afternoon/Evening.

3. A well-designed itinerary should include transportation arrangements, accommodation and key attractions, all organized in a proper chronological sequence to ensure a smooth travel experience, Pay special attention to the restrictions on the opening hours of attractions (openTimeCalendar field) to avoid scheduling visits during times when the attractions are closed or not allowing entry. 

4. For transportation, prioritize choosing main stations and pay special attention to the timing of transportation arrangements, ensuring they are scheduled within intended time period. Avoid mismatches such as planning morning departures for afternoon or evening time period.

5. For accommodation, find the most suitable hotel from the most suitable hotel area. One city only need one hotel and do not change hotel in the same city.

6. The hotel should be arranged at the end of the day, and hotel arrangements should be indicated every night, except on the last day.

7. It is especially important to pay attention to the time requirements for attractions and transportation to avoid time conflicts, which could lead to an unreasonable or unachievable itinerary.

8. Ensure that the travel schedule and physical exertion are moderate, and avoid arranging too many activities in the same time period. 

9. Transportation for the outbound and return trips is required, and be careful not to mix up the outbound and return trips.

10. If the user's travel duration is short (less than 2 days), it is recommended to focus on visiting the core attractions within the city.

\vspace{1em}

[Reference Data Rules]

Provided reference data may include:

    Attractions: poi (id: poiId, name: poiName)
    
    Hotels: hotel (id: hotelid, name: hotelname)
    
    Transportations: train/flight/bus/driving/ferry/ship (id: planid, name: trainNo/flightNo/shipName)
    
you must select hotels and transportation from reference data. But attractions are allowed to use external resources if more suitable.

\vspace{1em}

[Complete Plan JSON Specification]

1. Extract exact name, type, and id when using reference data:

    For attractions type: `poi', name: poiName, id: poiId. 
    
    For hotels type: `hotel', name: hotelname, id: hotelid; 
    
    For transportations type: `transportation', name: trainNo/flightNo, id: planid.

2. Attractions which is not in the reference data are allowed, but must set ``id'': ``'' and when mentioning the attraction in the description, use the format **attraction**. transportation and hotel must be chosen from the reference data.

3. for the items has the same id, just output one item, do not repeat items with different name but same id.

4. the Json format is as follows:

\{

\ \ ``itineraryName'': ``itinerary name like: 3 Days's Travel Itinerary: Shanghai to Beijing'',

\ \ ``recommendReason'': ``the reason why this itinerary is recommended, and make user feel that this itinerary is very suitable for him/her requirements. recommend reason should be no more than 50 words'',
 
\ \ ``dayInfos'': 

\ \  \  [

 \ \ \ \    \{
    
 \ \ \ \   \ \   ``day'': ``the order of days,a integer number starting from 1'',
        
\ \ \ \   \ \   ``scheduleTitle'': ``today's schedule title'',
        
\ \ \ \    \ \    ``scheduleDetail'': 
        
\ \ \ \     \ \         [
                        
\ \ \ \ \ \    \ \   \{ 
                            
\ \ \ \ \ \   \ \        ``period'': ``the time period when the schedule begins, must choose one from Morning/Afternoon/Evening(Capitalized Initial Letter)'',

\ \ \ \ \ \     ``description'':  ``Mention all attractions/hotels/transportations using the specified markdown syntax: For attractions and hotels,you should point out names and ids, use the format   \verb|**[PoiName](poiId)**| or \verb|**[HotelName](hotelId)**|, if the attractions is not in the reference data, use the format \verb|**[PoiName]**|.''
                                               
\ \ \ \ \ \     \ \    \ \    \ \   ``detailList'':  
                                
\ \ \ \ \ \     \ \      \ \  \ \       [
                                                
\ \ \ \ \ \  \ \        \ \    \ \  \ \    \ \      \{
                                                    
\ \ \ \ \ \  \ \         \ \       \ \  \ \    \ \      \ \      ``type'': ``transportation/poi/hotel'',
                                                        
\ \ \ \ \ \  \ \      \ \     \ \  \ \    \ \     \ \            ``id'': ``planid/poiId/hotelid'',
                                                        
\ \ \ \ \ \  \ \        \ \     \ \  \ \    \ \    \ \      ``name'': ``trainNo/flightNo/poiName/hotelname''
                                                        
\ \ \ \ \ \  \ \     \ \      \ \    \ \    \ \         \}
                                                    
\ \ \ \ \ \           \ \      \ \    \ \        ]
                                                
\ \ \ \ \ \       \     \      \}
                            
\ \ \ \      \ \    ]
                        
\ \   \ \    \}
    
\ ],

\  ``tips'': 

\ \  \{

\ \ \ \   ``title'': ``tips title'',

\ \ \ \    ``info'': ``the tips's total content should be within 50 words''

\ \    \}
    
\}

\vspace{1em}

******* Example **********

Input: I want to travel to Beijing, departing from Shanghai, for 3 days. Please design a detailed travel itinerary.

You may call FeasibleEnquiry like FeasibleEnquiry[\{
\ \ \ \ "day": 1,

\ \ \ \ "scheduleTitle": "Arrivel in Beijing and begins the tour with Beijing Zoo",

\ \ \ \ "scheduleDetail": [

\ \ \ \ \ \  \{

\ \ \ \ \ \ \ \ "period": "Morning",

\ \ \ \ \ \ \ \  "description": "Take the train **T110** from Shanghai to Beijing, the train duration is approximately 6 hours",

\ \ \ \ \ \ \ \   "detailList": [

\ \ \ \ \ \ \ \   \{

\ \ \ \ \ \ \ \ \ \    "type": "transportation",

\ \ \ \ \ \ \ \ \ \    "id": "1001",

\ \ \ \ \ \ \ \ \ \    "name": "T110"

\ \ \ \ \ \ \ \   \}

\ \ \ \ \ \ \ \   ]

\ \ \ \ \ \ \},

\ \ \ \ \ \  \{

\ \ \ \ \ \ \ \   "period": "Afternoon",

\ \ \ \ \ \ \ \   "description": "Exploring  \verb|**[Beijing Zoo](0001)**| and experience the vitality of the wildlife.",

\ \ \ \ \ \ \ \  "detailList": [

\ \ \ \ \ \ \ \ \ \    \{

\ \ \ \ \ \ \ \ \ \     "type": "poi",

\ \ \ \ \ \ \ \ \ \      "id": "0001",

\ \ \ \ \ \ \ \ \ \     "name": "Beijing Zoo"

\ \ \ \ \ \ \ \   \}

\ \ \ \ \ \ \ \   ]

\ \ \ \ \ \  \},

\ \ \ \ \ \  \{

\ \ \ \ \ \ \ \   "period": "Evening",

\ \ \ \ \ \ \ \   "description": "Stay in the Tiananmen Square Area for its proximity to magor attractions. Recommend \verb|**[Beijing Tiantan Manssion Hotel](0002)**| for its convenient transportation",

\ \ \ \ \ \ \ \   "detailList": [

\ \ \ \ \ \ \ \ \    \{

\ \ \ \ \ \ \ \ \ \   "type": "hotel",

\ \ \ \ \ \ \ \ \ \    "id": "0002",

\ \ \ \ \ \ \ \ \ \   "name": "Beijing Tiantan Manssion Hotel"

\ \ \ \ \ \ \ \ \  \}

\ \ \ \ \ \ \ \  ]

\ \ \ \ \ \   \}

\ \ \ \  ]

\ \ \}

]

You may call Finish like Finish[\{

\ \ "itineraryName": "3 Days's Travel Itinerary: Shanghai to Beijing",

\ \  "recommendReason": "Based on your request, with just three days, traveling from Shanghai to Beijing allows you to experience both cultural landmarks and modern attractions. I recommend this itinerary to explore Beijing's highlights at a comfortable pace.",

\ \  "dayInfos": [

\ \ \ \ "day": 1,

\ \ \ \ "scheduleTitle": "Arrivel in Beijing and begins the tour with Beijing Zoo",

\ \ \ \ "scheduleDetail": [

\ \ \ \ \ \  \{

\ \ \ \ \ \ \ \ "period": "Morning",

\ \ \ \ \ \ \ \  "description": "Take the train **T110** from Shanghai to Beijing, the train duration is approximately 6 hours",

\ \ \ \ \ \ \ \   "detailList": [

\ \ \ \ \ \ \ \   \{

\ \ \ \ \ \ \ \ \ \    "type": "transportation",

\ \ \ \ \ \ \ \ \ \    "id": "1001",

\ \ \ \ \ \ \ \ \ \    "name": "T110"

\ \ \ \ \ \ \ \   \}

\ \ \ \ \ \ \ \   ]

\ \ \ \ \ \ \},

\ \ \ \ \ \  \{

\ \ \ \ \ \ \ \   "period": "Afternoon",

\ \ \ \ \ \ \ \   "description": "Exploring  \verb|**[Beijing Zoo](0001)**| and experience the vitality of the wildlife.",

\ \ \ \ \ \ \ \  "detailList": [

\ \ \ \ \ \ \ \ \ \    \{

\ \ \ \ \ \ \ \ \ \     "type": "poi",

\ \ \ \ \ \ \ \ \ \      "id": "0001",

\ \ \ \ \ \ \ \ \ \     "name": "Beijing Zoo"

\ \ \ \ \ \ \ \   \}

\ \ \ \ \ \ \ \   ]

\ \ \ \ \ \  \},

\ \ \ \ \ \  \{

\ \ \ \ \ \ \ \   "period": "Evening",

\ \ \ \ \ \ \ \   "description": "Stay in the Tiananmen Square Area for its proximity to magor attractions. Recommend \verb|**[Beijing Tiantan Manssion Hotel](0002)**| for its convenient transportation",

\ \ \ \ \ \ \ \   "detailList": [

\ \ \ \ \ \ \ \ \    \{

\ \ \ \ \ \ \ \ \ \   "type": "hotel",

\ \ \ \ \ \ \ \ \ \    "id": "0002",

\ \ \ \ \ \ \ \ \ \   "name": "Beijing Tiantan Manssion Hotel"

\ \ \ \ \ \ \ \ \  \}

\ \ \ \ \ \ \ \  ]

\ \ \ \ \ \   \}

\ \ \ \  ]

\ \ \},

\ \  \{

\ \ \ \  "day": 2,

\ \ \ \     "scheduleTitle": "Explore Universal Beijing Resort and rediscover the joys of childhood ",

\ \ \ \     "scheduleDetail": [

\ \ \ \ \ \     \{

\ \ \ \ \ \  \ \   "period": "Morning",

\ \ \ \ \ \  \ \    "description": "After breakfast, take bus to  \verb|**[Universal Beijing Resort](0003)**| which is a large theme park resort located in Tongzhou District, including the Universal Studios Beijing theme park, two resort hotels, and a comprehensive commercial area",

\ \ \ \ \ \  \ \    "detailList": [

\ \ \ \ \ \ \ \ \ \      \{

\ \ \ \ \ \ \ \ \ \         "type": "poi",

\ \ \ \ \ \ \ \ \ \          "id": "0003",

\ \ \ \ \ \ \ \ \ \          "name": "Universal Beijing Resort"

\ \ \ \ \ \ \ \ \ \     \}

\ \ \ \ \ \  \ \    ]

\ \ \ \ \ \    \}

\ \ \ \    ]

\ \ \ \   \},

\ \ \ \   \{

\ \ \ \ \ \   "day": 3,

\ \ \ \ \ \   "scheduleTitle": "Cultural Immersion and Departure",

\ \ \ \ \ \  "scheduleDetail": [

\ \ \ \ \ \  \   \{

\ \ \ \ \ \  \ \     "period": "Morning",

\ \ \ \ \ \  \ \    "description": "Visit \verb|**[temple of Heaven](0004)**|: Explore this UNESCO Wold Heritage Site, where emperors prayed for good harvests",

\ \ \ \ \ \  \ \     "detailList": [

\ \ \ \ \ \  \ \ \       \{

\ \ \ \ \ \  \ \ \ \       "type": "poi",

\ \ \ \ \ \  \ \ \ \        "id": "0004",

\ \ \ \ \ \  \ \ \ \       "name": "temple of Heaven"

\ \ \ \ \ \  \ \ \     \}

\ \ \ \ \ \  \ \ \    ]

\ \ \ \ \ \  \ \   \}

\ \ \ \ \ \    ]

\ \ \ \   \}

\ \  ],

\ \  "tips": \{

\ \ \ \    "title": "Tips for Enhanced Travel Experience",

\ \ \ \  "info": [

\ \ \ \ \ \     "Use a private car or join a guided tour for the Universal Beijing Resort",

\ \ \ \ \ \    "Book tickets in advance to secure your entry",

\ \ \ \ \ \     "Pack comfortable walkina shoes"

\ \ \ \   ]

\ \   \}

\}

]

******* End of Example **********

You must call Finish to indicate you have completed the task. Each action must call exactly one function; do not call multiple functions at the same time. Please make sure call Finish before 5 rounds of dialogue.

\vspace{1em}

\vspace{1em}
[References]

[transportation arrangements]

\{transportation\_information\}
\vspace{1em}

[attractions reference information]

\{attraction\_information\}
\vspace{1em}

[hotels reference information]

\{hotel\_information\}

\vspace{1em}

[User Query]

\{user\_request\}

\end{tcolorbox}

\subsection{Point-wise evaluation Prompt}
\label{point_wise}
\begin{tcolorbox}
[title=     POINT\_WISE\_EVALUATION\_PROMPT,colback=blue!10,colframe=blue!50!black,arc=1mm,boxrule=1pt,left=1mm,right=1mm,top=1mm,bottom=1mm, breakable]
\small
You are a travel itinerary quality reviewer. Please rate a single itinerary based on the following criteria (0-100), without introducing external information or speculation.

\vspace{1em}

[Evaluation Criteria] (in order of priority from high to low)

\vspace{1em}

1. Format and Facts (hard constraints, severe violations directly Inferior)

\begin{itemize}[label={}, leftmargin=1em,itemsep=0em, parsep=0em]
\item Response structure: The output must strictly follow the requested schema. Missing or misplaced elements are non-compliant.
\item Information verification: Transportation/hotels/attractions must come from the given text; introducing external facts or conjecture is treated as hallucination and deemed invalid.
\item Information accuracy: Details such as names/times are consistent;
\item Information relevance: Description matches corresponding attractions/events.

\end{itemize}

2. Common Sense and Feasibility (hard constraints)

\begin{itemize}[label={}, leftmargin=1em,itemsep=0em, parsep=0em]
\item   Complete information: Each destination must include necessary accommodation, essential transportation, and key activities to ensure executability.
\item   Correct time sequence: Activities must be listed in temporal order; days cannot backtrack, and intra-day sequences must be non-decreasing in time.
\item   Location consistency: A traveler cannot be in multiple cities/locations simultaneously; any change of location must be justified by an explicit transport step.
\item   Feasible operating hours: Visits must occur within confirmed opening hours; closed days/times invalidate scheduled activities.
\item   Transportation block: No activities scheduled during transport intervals;
\item   Early transportation rule: If departure time is before 10 AM, no earlier activities scheduled that day;
\item   Transportation continuity: Smooth movement between cities/attractions, no repeated backtracking.

\end{itemize}

3. Soft Constraints

\begin{itemize}[label={}, leftmargin=1em,itemsep=0em, parsep=0em]
\item   Moderate pace density: Daily pacing should be balanced—neither overpacked nor overly sparse—with reasonable buffers for transition and rest.
\item   Hotel Consistency: Within the same city, prefer a single hotel to minimize check-in/out overhead and travel friction.
\item   Daytime Utilization: Prioritize activities during daylight; reserve evenings for appropriate activities or rest, avoiding unproductive daytime gaps.
\item   Unique Attractions: Avoid repeated visits to the same (or effectively identical) attractions.
\item   Location Clustering: Group nearby attractions to reduce transit time and improve route efficiency.
\item   Iconic Landmarks: When feasible, include representative, must-see landmarks to improve coverage and recognizability.
\item   Attraction Diversity: Maintain variety across categories (e.g., cultural, natural, museums, landmarks) to avoid monotony.
\end{itemize}

4. Preference Matching (only considered if preferences appear in the text, otherwise treated neutrally)

\begin{itemize}[label={}, leftmargin=1em,itemsep=0em, parsep=0em]
\item  Budget Preference: Select hotels/activities aligned with the stated budget profile (e.g., premium, budget-conscious, value-oriented).
\item  Pacing Preference: Match the requested pacing (relaxed, moderate, compact) by adjusting daily activity counts and durations.
\item  Attraction Prioritization: Prioritize categories explicitly favored by the user and ensure requested items are covered.
\item  Physical Effort Preference: Align walking distance and intensity with the specified level (light, moderate, strenuous), managing elevation and high-exertion activities accordingly.
\item  User Request Fulfillment: Satisfy explicit user constraints (e.g., must-visit/avoid, time windows, ordering). If no preferences are stated, no penalty or credit is applied.

\end{itemize}

\vspace{1em}

[Scoring Approach]

First apply compliance deductions based on hard constraints, then provide a total score considering soft constraints and preference matching.

\vspace{1em}

[Scoring Anchors]

90-100: Comprehensive, factually accurate, highly actionable, well-paced, and strongly aligned with preferences.

70-85: Largely complete, with occasional minor flaws that do not impede execution or user experience.

50-65: Moderate quality; contains several issues but remains executable.

30-45: Significant flaws (e.g., temporal/spatial conflicts, missing elements) requiring substantial revision.

0-25: Numerous severe issues or a large amount of fabricated/irrelevant information; largely unusable/inactionable.

\vspace{1em}

[Output Requirement]

Strictly output `score' (0-100), no explanations or additional text.

\vspace{1em}

[User Query]

\{user\_request\}

\vspace{1em}

[Itinerary]

\{itinerary\}

\end{tcolorbox}

\subsection{Pair-wise evaluation Prompt}
\label{pair_wise}
\begin{tcolorbox}
[title=     PAIR\_WISE\_EVALUATION\_PROMPT,colback=blue!10,colframe=blue!50!black,arc=1mm,boxrule=1pt,left=1mm,right=1mm,top=1mm,bottom=1mm, breakable]
\small
You are a travel itinerary quality reviewer. Your task is to compare two candidate itineraries under strict evaluation criteria, without introducing external information or speculation.

\vspace{1em}

[Evaluation Criteria] (in order of priority from high to low)


1. Format and Facts (hard constraints, severe violations directly Inferior)
\begin{itemize}[label={}, leftmargin=1em,itemsep=0em, parsep=0em]
\item Response structure: The output must strictly follow the requested schema. Missing or misplaced elements are non-compliant.
\item Information verification: Transportation/hotels/attractions must come from the given text; introducing external facts or conjecture is treated as hallucination and deemed invalid.
\item Information accuracy: Details such as names/times are consistent;
\item Information relevance: Description matches corresponding attractions/events.

\end{itemize}

2. Common Sense and Feasibility (hard constraints)
\begin{itemize}[label={}, leftmargin=1em,itemsep=0em, parsep=0em]
\item   Complete information: Each destination must include necessary accommodation, essential transportation, and key activities to ensure executability.
\item   Correct time sequence: Activities must be listed in temporal order; days cannot backtrack, and intra-day sequences must be non-decreasing in time.
\item   Location consistency: A traveler cannot be in multiple cities/locations simultaneously; any change of location must be justified by an explicit transport step.
\item   Feasible operating hours: Visits must occur within confirmed opening hours; closed days/times invalidate scheduled activities.
\item   Transportation block: No activities scheduled during transport intervals;
\item   Early transportation rule: If departure time is before 10 AM, no earlier activities scheduled that day;
\item   Transportation continuity: Smooth movement between cities/attractions, no repeated backtracking.

\end{itemize}

3. Soft Constraints
\begin{itemize}[label={}, leftmargin=1em,itemsep=0em, parsep=0em]
\item   Moderate pace density: Daily pacing should be balanced—neither overpacked nor overly sparse—with reasonable buffers for transition and rest.
\item   Hotel Consistency: Within the same city, prefer a single hotel to minimize check-in/out overhead and travel friction.
\item   Daytime Utilization: Prioritize activities during daylight; reserve evenings for appropriate activities or rest, avoiding unproductive daytime gaps.
\item   Unique Attractions: Avoid repeated visits to the same (or effectively identical) attractions.
\item   Location Clustering: Group nearby attractions to reduce transit time and improve route efficiency.
\item   Iconic Landmarks: When feasible, include representative, must-see landmarks to improve coverage and recognizability.
\item   Attraction Diversity: Maintain variety across categories (e.g., cultural, natural, museums, landmarks) to avoid monotony.
\end{itemize}

4. Preference Matching (only considered if preferences appear in the text, otherwise treated neutrally)
\begin{itemize}[label={}, leftmargin=1em,itemsep=0em, parsep=0em]
\item  Budget Preference: Select hotels/activities aligned with the stated budget profile (e.g., premium, budget-conscious, value-oriented).
\item  Pacing Preference: Match the requested pacing (relaxed, moderate, compact) by adjusting daily activity counts and durations.
\item  Attraction Prioritization: Prioritize categories explicitly favored by the user and ensure requested items are covered.
\item  Physical Effort Preference: Align walking distance and intensity with the specified level (light, moderate, strenuous).
\item  User Request Fulfillment: Satisfy explicit user constraints (e.g., must-visit/avoid, time windows, ordering). 

\end{itemize}

\vspace{1em}

[Decision]

First pay attention to hard constraints, severe violations are inferior; if both comply, then compare soft constraints and preference matching. If difficult to distinguish, choose the clearer and more executable one.

\vspace{1em}

[Output Requirement]

Only output ``Route A'' or ``Route B'', no other characters or explanations.

\vspace{1em}

[User Query]

\{user\_request\}

\vspace{1em}

[Route A]

\{route\_A\}

\vspace{1em}

[Route B]

\{route\_B\}

\end{tcolorbox}

\end{document}